\documentclass[]{bytedance_seed}
\usepackage[T1]{fontenc}
\usepackage[toc,page,header]{appendix}

\usepackage{wrapfig}
\usepackage{multirow}
\usepackage{makecell}
\usepackage{xcolor}
\usepackage{colortbl}

\newcommand{\paperfullname}{\textbf{Depth Anything 3}\xspace}

\usepackage{amsmath,amsfonts,bm}

\usepackage{xspace}

\newcommand{\picodataset}{HiRoom}
\newcommand{\shortname}{DA3 }
\newcommand{\longname}{Depth Anything 3 }
\newcommand{\giantmodel}{DA3-Gaint\xspace}
\newcommand{\largemodel}{DA3-Large\xspace}
\newcommand{\pcd}{{\fontfamily{ppl}\selectfont \textbf{pcd}}\xspace}
\newcommand{\depth}{{\fontfamily{ppl}\selectfont \textbf{depth}}\xspace}
\newcommand{\ray}{{\fontfamily{ppl}\selectfont \textbf{ray}}\xspace}
\newcommand{\cam}{{\fontfamily{ppl}\selectfont \textbf{cam}}\xspace}
\newcommand{\ie}{\textit{i.e.}\xspace}
\newcommand{\eg}{\textit{e.g.}\xspace}

\newcommand{\markedblanklines}[2][\textbullet]{
  \par
  \begingroup
  \setlength{\parskip}{0pt}
  \setlength{\parindent}{0pt}
  \count0=0
  \loop
    \ifnum\count0<#2
      #1\hspace{1em}\rule{0pt}{\baselineskip}\par
      \advance\count0 by 1
  \repeat
  \endgroup
}

\def\eqref#1{equation~\ref{#1}}

\def\1{\bm{1}}

\DeclareMathAlphabet{\mathsfit}{\encodingdefault}{\sfdefault}{m}{sl}
\SetMathAlphabet{\mathsfit}{bold}{\encodingdefault}{\sfdefault}{bx}{n}

\newcommand{\inputset}{\mathcal{I}}
\newcommand{\inputimage}{\mathbf{I}}
\newcommand{\depthmap}{\mathbf{D}}
\newcommand{\camerarot}{\mathbf{R}}
\newcommand{\cameratrans}{\mathbf{t}}
\newcommand{\cameraint}{\mathbf{K}}
\newcommand{\cameravec}{\mathbf{v}}
\newcommand{\rotquat}{\mathbf{q}}
\newcommand{\fovparam}{\mathbf{f}}
\newcommand{\pixelcoord}{\mathbf{p}}
\newcommand{\worldpoint}{\mathbf{P}}
\newcommand{\cameraray}{\mathbf{r}}
\newcommand{\rayorigin}{\mathbf{t}}
\newcommand{\raydir}{\mathbf{d}}
\newcommand{\raymap}{\mathbf{M}}
\newcommand{\camerahead}{\mathcal{D}_C}

\newcommand{\transformerlayers}{L}
\newcommand{\singleviewlayers}{L_\text{s}}
\newcommand{\globallayers}{L_\text{g}}
\newcommand{\cameraencoder}{\mathcal{E}_{c}}
\newcommand{\cameratoken}{\mathbf{c}}
\newcommand{\learnabletoken}{\mathbf{c}_l}

\newcommand{\model}{\mathcal{F}_\theta}
\newcommand{\preddepth}{\hat{\mathbf{D}}}
\newcommand{\predray}{\hat{\mathbf{R}}}
\newcommand{\predcamera}{\hat{\mathbf{c}}}
\newcommand{\totalloss}{\mathcal{L}}
\newcommand{\depthloss}{\mathcal{L}_{D}}
\newcommand{\rayloss}{\mathcal{L}_{M}}
\newcommand{\pointloss}{\mathcal{L}_{P}}
\newcommand{\cameraloss}{\mathcal{L}_{C}}
\newcommand{\gradloss}{\mathcal{L}_\text{grad}}
\newcommand{\gradweight}{\alpha}
\newcommand{\cameraweight}{\beta}
\newcommand{\gradx}{\nabla_x}
\newcommand{\grady}{\nabla_y}
\newcommand{\mseloss}{\mathcal{L}_\mathrm{MSE}}
\newcommand{\lpipsloss}{\mathcal{L}_\mathrm{LPIPS}}

\definecolor{colorfirst}{rgb}{.866,.945, 0.831}
\definecolor{colorsecond}{rgb}{1, 0.98, 0.83}
\definecolor{colorthird}{rgb}{0.76, 0.87, 0.92}
\definecolor{colorcite}{rgb}{0.212, 0.490, 0.741}

\newcommand{\cellfirst}{\cellcolor{colorfirst}}
\newcommand{\cellsecond}{\cellcolor{colorsecond}}
\newcommand{\cellthird}{\cellcolor{colorthird}}

\newcommand{\textfirst}{\colorbox{colorfirst}}
\newcommand{\secondtext}{\colorbox{colorsecond}}
\newcommand{\thirdtext}{\colorbox{colorthird}}

\usepackage{cleveref}
\crefname{figure}{Fig.}{Figs.}
\crefname{table}{Tab.}{Tabs.}
\crefname{equation}{Eq.}{Eqs.}
\crefname{section}{Sec.}{Secs.}

\title{Depth Anything 3: \\ Recovering the Visual Space from Any Views}

\author[*]{Haotong Lin}
\author[*]{Sili Chen}
\author[*]{Jun Hao Liew}
\author[*]{Donny Y. Chen}
\author{Zhenyu Li}
\author{Guang Shi}
\author{Jiashi Feng}
\author[*,\dagger]{Bingyi Kang}

\affiliation[]{ByteDance Seed}

\contribution[\dagger]{Project Lead}
\contribution[*]{Equal Contribution}

\abstract{
We present Depth Anything 3 (DA3), a model that predicts spatially consistent geometry from an arbitrary number of visual inputs, with or without known camera poses. 
In pursuit of minimal modeling, DA3 yields two key insights:
a single plain transformer (\eg, vanilla DINO encoder) is sufficient as a backbone without architectural specialization, and a singular depth-ray prediction target obviates the need for complex multi-task learning. Through our teacher-student training paradigm, the model achieves a level of detail and generalization on par with Depth Anything 2 (\textit{DA2}).
We establish a new visual geometry benchmark covering camera pose estimation, any-view geometry and visual rendering. On this benchmark, DA3 sets a new state-of-the-art across all tasks, surpassing prior SOTA \textit{VGGT} by an average of 35.7\% in camera pose accuracy and 23.6\% in geometric accuracy. Moreover, it outperforms \textit{DA2} in monocular depth estimation. All models are trained exclusively on public academic datasets.
}

\correspondence{Bingyi Kang}

\checkdata[Project Page]{\url{depth-anything-3.github.io}}

\begin{document}

\maketitle

\begin{figure}[h]
    \centering
    \includegraphics[width=\linewidth]{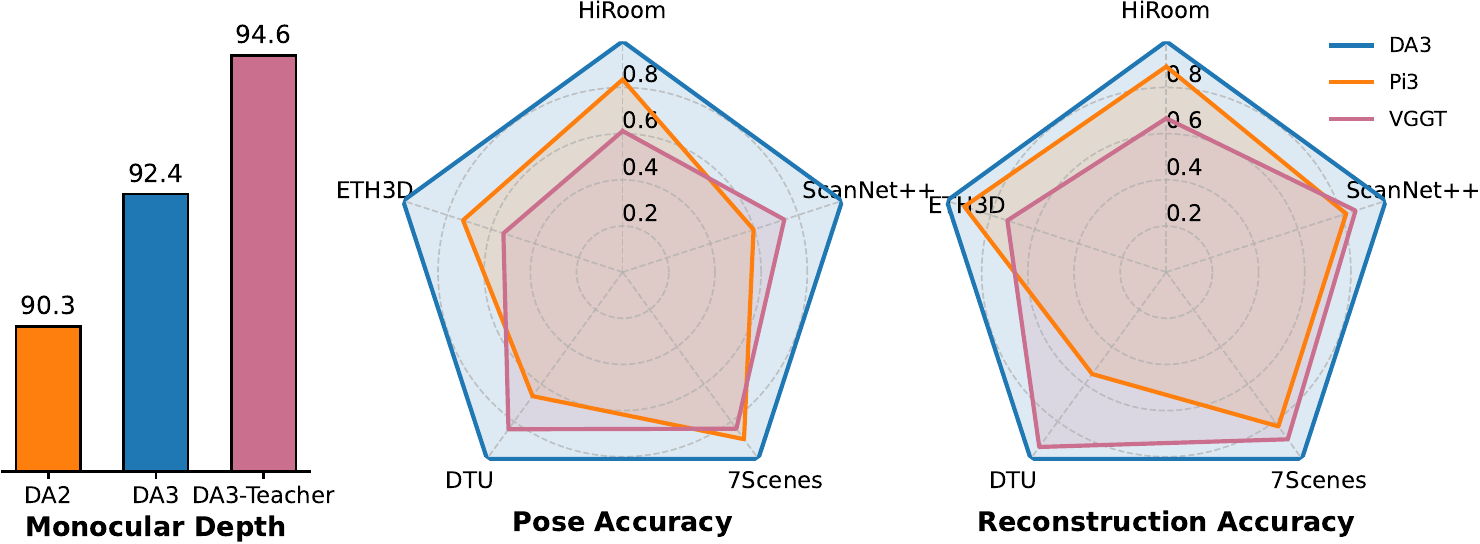}
    \label{fig:placeholder}
    \vspace{-3mm}
\end{figure}

\footnotetext[1]{``Depth Anything 3'' marks a new generation for the series, expanding from monocular to any-view inputs, built on our conviction that depth is the cornerstone of understanding the physical world.}

\newpage
\tableofcontents
\newpage

\begin{figure}[h]
    \vspace{0mm}
    \begin{center}
        \includegraphics[width=1\textwidth]{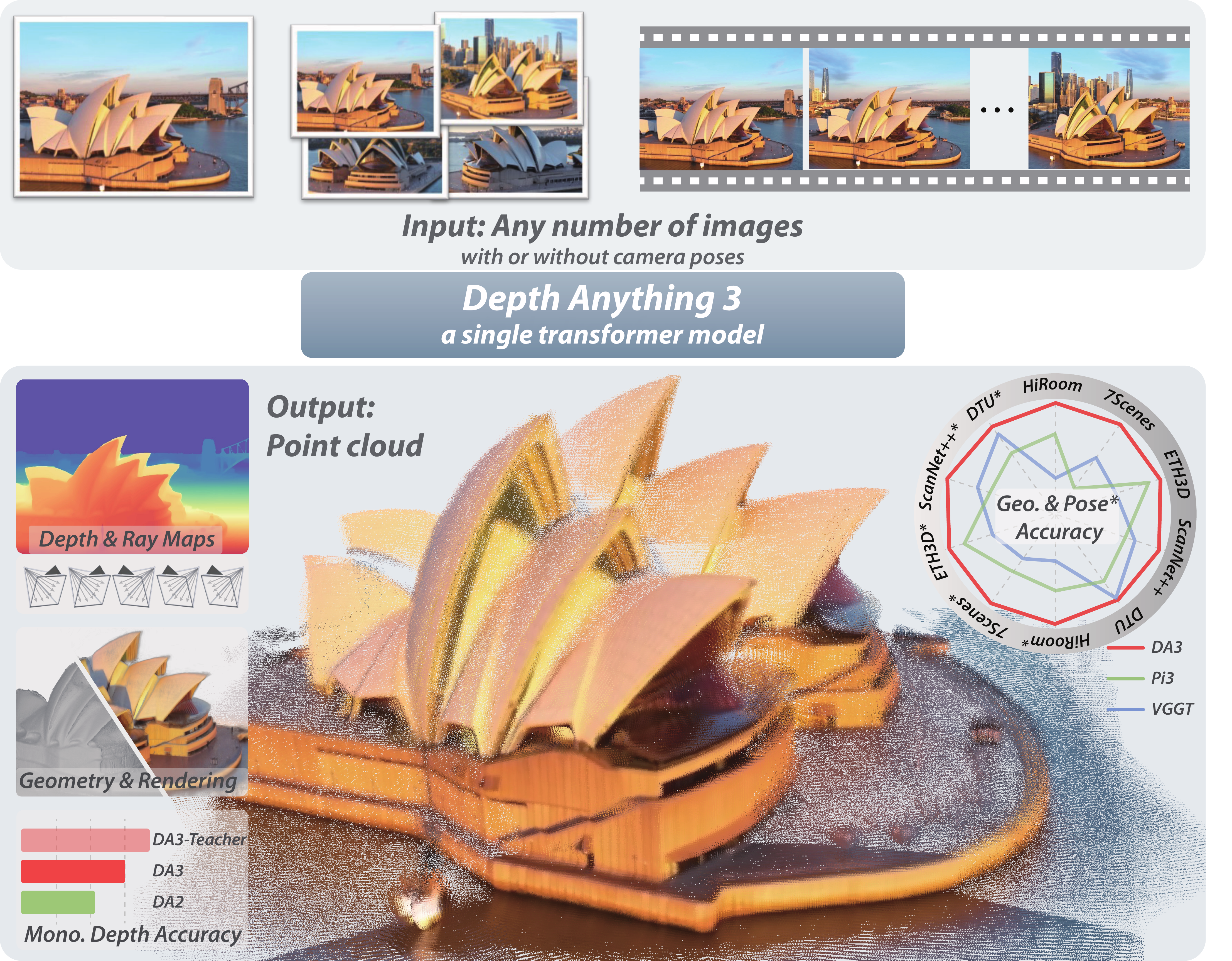}
    \caption{ 
    Given any number of images and optional camera poses, \paperfullname reconstructs the visual space, producing consistent depth and ray maps that can be fused into accurate point clouds, resulting in high-fidelity 3D Gaussians and geometry. 
    It significantlyoutperforms VGGT in multi-view geometry and pose accuracy; with monocular inputs, it also surpasses Depth Anything 2 while matching its detail and robustness.
    }
    \label{fig:teaser}
    \end{center}
\end{figure}\bigbreak

\section{Introduction}

The ability to perceive and understand 3D spatial information from visual input is a cornerstone of human spatial intelligence~\citep{arterberry2000perception} and a critical requirement for applications like robotics and mixed reality. This fundamental capability has inspired a wide array of 3D vision tasks, including Monocular Depth Estimation~\citep{eigen2014depth}, Structure from Motion~\citep{snavely2006photo}, Multi-View Stereo~\citep{seitz2006comparison} and Simultaneous Localization and Mapping~\citep{mur2015orb}. Despite the strong conceptual overlap between these tasks—often differing by only a single factor, such as the number of input views—the prevailing paradigm has been to develop highly specialized models for each one. While recent efforts~\citep{wang2024dust3r,wang2025vggt} have explored unified models to address multiple tasks simultaneously, 
they typically suffer from key limitations: they often rely on complex, bespoke architectures, are trained via joint optimization over tasks from scratch, and consequently cannot effectively leverage large-scale pretrained models.

In this work, we step back from established 3D task definitions and return to a more fundamental goal inspired by human spatial intelligence: recovering 3D structure from arbitrary visual inputs, be it a single image, multiple views of a scene, or a video stream. Forsaking intricate architectural engineering, we pursue a minimal modeling strategy guided by two central questions. First, \textit{is there \textbf{a minimal set of prediction targets}, or is joint modeling across numerous 3D tasks necessary?} Second, \textit{can a \textbf{single plain transformer} suffice for this objective}?
Our work provides an affirmative answer to both. We present \paperfullname, a single transformer model trained exclusively for joint \textbf{any-view depth and pose estimation} via a specially chosen ray representation. We demonstrate that this minimal approach is sufficient to reconstruct the visual space from any number of images, with or without known camera poses.

\paperfullname formulates the above geometric reconstruction target as a dense prediction task. For a given set of $N$ input images, the model is trained to output $N$ corresponding depth maps and ray maps, each pixel-aligned with its respective input. The architecture to achieve this begins with a standard pretrained vision transformer (e.g., ~\citealt{oquab2023dinov2}), as its backbone, leveraging its powerful feature extraction capabilities. To handle arbitrary view counts, we introduce a key modification: an input-adaptive cross-view self-attention mechanism. This module dynamically rearranges tokens during the forward pass in selected layers, enabling efficient information exchange across all views. For the final prediction, we propose a new dual DPT head designed to jointly outputs both depth and ray values, by processing the same set of features with distinct fusion parameters. To enhance flexibility, the model can optionally incorporate known camera poses via a simple camera encoder, allowing it to adapt to various practical settings. This overall design results in a clean and scalable architecture that directly inherits the scaling properties of its pretrained backbone.

We train Depth Anything 3 via a teacher-student paradigm to unify diverse training data, which is necessary for a generalist model. Our data sources include varied formats like real-world depth camera captures (e.g., \citealt{baruch2021arkitscenes}), 3D reconstruction (e.g., \citealt{reizenstein2021common}), and synthetic data, where real-world depth may be of poor quality (\cref{fig:poordataset}).
To resolve this, we adopt a pseudo-labeling strategy inspired by prior works~\citep{yang2024depth,yang2024depthv2}. Specifically, we train a powerful teacher monocular depth model on synthetic data to generate dense, high-quality pseudo-depth for all real-world data. Crucially, to preserve geometric integrity, we align these dense pseudo-depth maps with the original sparse or noisy depth. This approach proved remarkably effective, significantly enhancing label detail and completeness without sacrificing the geometric accuracy.

To better evaluate our model and track progress in the field, we establish a comprehensive benchmark for assessing geometry and pose accuracy.
The benchmark comprises 5 distinct datasets, totaling over 89 scenes, ranging from object-level to indoor and outdoor environments.
By directly evaluating pose accuracy across scenes and fusing the predicted pose and depth into a 3D point cloud for accuracy assessment, the benchmark faithfully measures the pose and depth accuracy of visual geometry estimators.
Experiments show that our model achieves state-of-the-art performance on 18 out of 20 settings.
Moreover, on standard monocular benchmarks, our model outperforms Depth Anything 2~\citep{yang2024depthv2}.

To further demonstrate the fundamental capability of \longname in advancing other 3D vision tasks, we introduce a challenging benchmark for feed-forward novel view synthesis (FF-NVS), comprising over 160 scenes.
We adhere to the minimal modeling strategy and fine-tune our model with an additional DPT head to predict pixel-aligned 3D Gaussian parameters. 
Extensive experiments
yield two key findings: 1) fine-tuning a geometry foundation model for NVS substantially outperforms highly specialized task-specific models~\citep{xu2025depthsplat}; 2) enhanced geometric reconstruction capability directly correlates with improved FF-NVS performance, establishing \longname as the optimal backbone for this task.

\section{Related Work}

\paragraph{Multi-view visual geometry estimation.}
Traditional systems~\citep{schoenberger2016sfm, schoenberger2016mvs} decompose reconstruction into feature detection and matching, robust relative pose estimation, incremental or global SfM with bundle adjustment, and dense multi-view stereo for per-view depth and fused point clouds. These methods remain strong on well-textured scenes, but their modularity and brittle correspondences complicate robustness under low texture, specularities, or large viewpoint changes.
Early learning methods injected robustness at the component level: 
learned detectors~\citep{detone2018superpoint}, descriptors for matching~\citep{dusmanu2019d2}, and differentiable optimization layers that expose pose/depth updates to gradient flow~\citep{he2024detector,guo2025multi,pan2024global}. 
On the dense side, cost-volume networks~\citep{yao2018mvsnet,xu2023iterative} for MVS replaced hand-crafted regularization with 3D CNNs, improving depth accuracy especially at large baselines and thin structures compared with classical PatchMatch.
Early end-to-end approaches~\citep{teed2018deepv2d,wang2024vggsfm} moved beyond modular SfM/MVS pipelines by directly regressing camera poses and per-image depths from pairs of images. 
These approaches reduced engineering complexity and demonstrated the feasibility of learned joint depth pose estimation, but they often struggled with scalability, generalization, and handling arbitrary input cardinalities. 

A turning point came with DUSt3R~\citep{dust3r}, which leveraged transformers to directly predict point map between two views and compute both depth and relative pose in a purely feed-forward manner. This work laid the foundation for subsequent transformer-based methods aiming to unify multi-view geometry estimation at scale. Follow-up models extended this paradigm with multi-view inputs \citep{fast3r,cut3r,Mv-dust3r+,must3r}, video input~\citep{monst3r,cut3r,murai2025mast3r,deng2025vggtlong}, robust correspondence modeling ~\citep{mast3r}, camera parameter injection~\citep{jang2025pow3r,keetha2025mapanything}, large-scale SfM~\citep{deng2025sailrecon}, SLAM applications~\citep{maggio2025vggtslam}, and view synthesis with 3D Gaussians~\citep{flare,smart2024splatt3r,charatan2024pixelsplat,chen2024mvsplat,jiang2025anysplat,xu2025depthsplat}. Among these, \cite{wang2025vggt} push accuracy to a new level through large-scale training, a multi-stage architecture, and redundancy in design. In contrast, we focus on a minimal modeling strategy built around a single, simple transformer.

\paragraph{Monocular depth estimation.}
Early monocular depth estimation methods relied on fully supervised learning on single-domain datasets, which often produced models specialized to either indoor rooms~\citep{silberman2012indoor} or outdoor driving scenes~\citep{geiger2013vision}. These early deep models achieved good accuracy within their training domain but struggled to generalize to novel environments, highlighting the challenge of cross-domain depth prediction. 
Modern generalist approaches~\citep{yang2024depth,yang2024depthv2,wang2025moge,bochkovskii2024depth,yin2023metric3d,ke2024repurposing} exemplify this trend by leveraging massive multi-dataset training and advanced architectures like vision transformers~\citep{ranftl2021vision} or DiT~\citep{peebles2023scalable}.
Trained on millions of images, they learn broad visual cues and incorporate techniques such as affine-invariant depth normalization.
In contrast, our method is primarily designed for a unified visual geometry estimation task, yet it still demonstrates competitive monocular depth performance.

\paragraph{Feed-Forward Novel View Synthesis}

Novel view synthesis (NVS) has long been a core problem in computer vision and graphics~\cite{levoy1996light,heigl1999plenoptic,buehler2001unstructured}, and interest has increased with the rise of neural rendering~\cite{sitzmann2019scene,sitzmann2021light,mildenhall2020nerf,kerbl20233d,govindarajan2025radiant}. A particularly promising direction is feed-forward NVS, which produces 3D representations in a single pass through an image-to-3D network, avoiding tedious per-scene optimization. Early methods adopted NeRF as the underlying 3D representation~\cite{yu2021pixelnerf,chen2021mvsnerf,lin2022efficient,chen2025explicit,hong2024lrm,xu2024murf}, but recent work has largely shifted to 3DGS due to its explicit structure and real-time rendering. Representative approaches improve image-to-3D networks with geometry priors, \eg, epipolar attention~\cite{charatan2024pixelsplat}, cost volumes~\cite{chen2024mvsplat}, and depth priors~\cite{xu2024murf}. More recently, multi-view geometry foundation models~\cite{dust3r,Mv-dust3r+,fast3r,wang2025vggt} have been integrated to improve modeling capacity, particularly in pose-free settings, yet methods built upon such models are often evaluated by relying on a single chosen foundation model~\cite{smart2024splatt3r,ye2024no,jiang2025anysplat}. Here, we systematically benchmark the contribution of different geometry foundation models to NVS and propose strategies to better exploit them, enabling feed-forward 3DGS to handle both posed and pose-free inputs, variable numbers of views, and arbitrary resolutions.

\begin{figure}
    \begin{center}
        \includegraphics[width=0.95\textwidth]{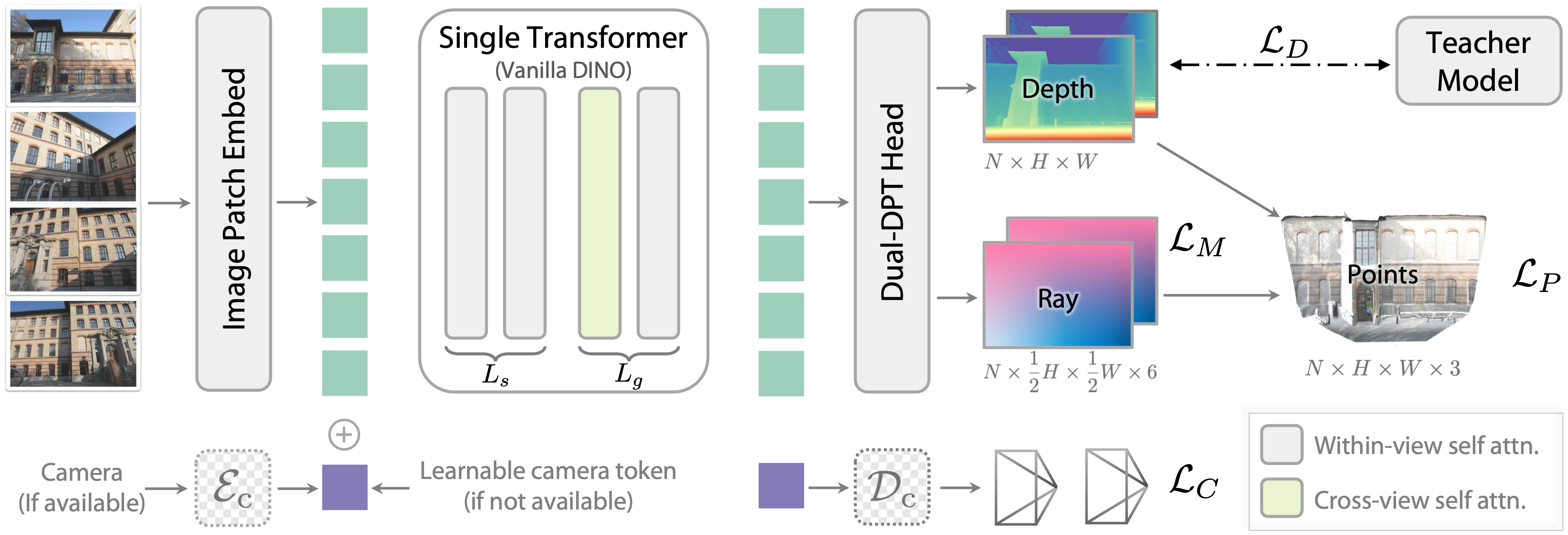}
        \caption{\textbf{Pipeline of Depth Anything 3.} 
        Depth Anything 3 employs a single transformer (vanilla DINOv2 model) without any architectural modifications. To enable cross-view reasoning, an input-adaptive cross-view self-attention mechanism is introduced. A dual-DPT head is used to predict depth and ray maps from visual tokens.
        Camera parameters, if available, are encoded as camera tokens and concatenated with patch tokens, participating in all attention operations.  
        }
        \label{fig:pipeline}
    \end{center}
\end{figure}

\section{Depth Anything 3}
\label{sec:method}

We tackle the recovery of consistent 3D geometry from diverse visual inputs—single image, multi-view collections, or videos—and optionally incorporate known camera poses when available.

\subsection{Formulation}
\label{sec:formulation}

We denote the input as $\inputset=\{\inputimage_i\}_{i=1}^{N_v}$ with each $\inputimage_i \in \mathbb{R}^{H \times W \times 3}$. For $N_v=1$ this is a monocular image, and for $N_v>1$ it represents a video or multi-view set. Each image has depth $\depthmap_i \in \mathbb{R}^{H \times W}$, camera extrinsics $\big[\camerarot_i \mid \cameratrans_i\big]$, and intrinsics $\cameraint_i$. The camera can also be represented as $\cameravec_i \in \mathbb{R}^9$ with translation $\cameratrans_i\in \mathbb{R}^3$, rotation quaternion $\rotquat_i\in \mathbb{R}^4$, and FOV parameters $\fovparam_i\in \mathbb{R}^2$. A pixel $\pixelcoord=(u,v,1)^\top$ projects to a 3D point $\worldpoint=(X,Y,Z,1)^\top$ by  
\begin{equation*}
    \worldpoint=\camerarot_i\big(\depthmap_i(u,v)\,\cameraint_i^{-1}\pixelcoord\big)+\cameratrans_i,
\end{equation*}  
through which the underlying 3D visual space can be faithfully recovered.

\paragraph{Depth-ray representation.} Predicting a valid rotation matrix $\mathbf{R}_i$ is challenging due to the orthogonality constraint. To avoid this, we represent camera pose implicitly with a per-pixel ray map, aligned with the input image and depth map. For each pixel $\pixelcoord$, the camera ray $\cameraray \in \mathbb{R}^6$ is defined by its origin $\rayorigin \in \mathbb{R}^3$ and direction $\raydir \in \mathbb{R}^3$: $\cameraray = (\rayorigin, \raydir)$. The direction is obtained by backprojecting $\pixelcoord$ into the camera frame and rotating it to the world frame: 
$
    \raydir = \camerarot \cameraint^{-1} \pixelcoord.
$
The dense ray map $\raymap \in \mathbb{R}^{H \times W \times 6}$ stores these parameters for all pixels. We do not normalize $\raydir$, so its magnitude preserves the projection scale. Thus, a 3D point in world coordinates is simply
$
    {\worldpoint} = {\rayorigin} + \depthmap(u, v) \cdot \raydir.
    \label{eq:depth_to_pcd}
$
This formulation enables consistent point cloud generation by combining predicted depth and ray maps through element-wise operations.

\label{sec:ray2pose}
\paragraph{Deriving Camera Parameters from the Ray Map.}
Given an input image $\inputimage \in \mathbb{R}^{H\times W \times 3}$, the corresponding ray map is denoted by $\raymap \in \mathbb{R}^{H \times W \times 6}$. This map comprises per-pixel ray origins, stored in the first three channels ($\raymap(:,:,:3)$), and ray directions, stored in the last three ($\raymap(:,:,3:)$). 

First, the camera center $\rayorigin_c$ is estimated by averaging the per-pixel ray origin vectors:
\begin{equation}
    \rayorigin_c = \frac{1}{H \times W} \sum_{h=1}^{H}\sum_{w=1}^W \raymap(h, w, :3).
\end{equation}

To estimate the rotation $\camerarot$ and intrinsics $\cameraint$, we formulate the problem as finding a homography $\mathbf{H}$. 
We begin by defining a  ``identity'' camera with an identity intrinsics matrix, $\cameraint_I = \mathbf{I}$. 
For a given pixel $\pixelcoord$, its corresponding ray direction in this canonical camera's coordinate system is simply $\mathbf{d}_I = \cameraint_I^{-1} \pixelcoord = \pixelcoord$. 
The transformation from this canonical ray $\mathbf{d}_I$ to the ray direction $\mathbf{d}_\text{cam}$ in the target camera's coordinate system is given by $\mathbf{d}_\text{cam} = \cameraint \camerarot \mathbf{d}_I$. 
This establishes a direct homography relationship, $\mathbf{H} = \cameraint \camerarot$, between the two sets of rays.
We can then solve for this homography by minimizing the geometric error between the transformed canonical rays and a set of pre-computed target rays, $\raymap(h,w,3:)$. This leads to the following optimization problem:
\begin{equation}
    \mathbf{H}^* = \arg\min_{||\mathbf{H}|| = 1}\sum_{h=1}^{H}\sum_{w=1}^W \left|\left| \mathbf{H} \pixelcoord_{h,w} \times \raymap(h,w,3:) \right|\right|.
\end{equation}
This is a standard least-squares problem that can be efficiently solved using the Direct Linear Transform (DLT) algorithm~\citep{abdel2015direct}. 
Once the optimal homography $\mathbf{H}^*$ is found, we recover the camera parameters. 
Since the intrinsic matrix $\cameraint$ is upper-triangular and the rotation matrix $\camerarot$ is orthonormal, we can uniquely decompose $\mathbf{H}^*$ using RQ decomposition to obtain $\cameraint$, $\camerarot$.

\paragraph{Minimal prediction targets.} Recent works aim to build unified models for diverse 3D tasks, often using multitask learning with different targets---for example, point maps alone~\citep{dust3r}, or redundant combinations of pose, local/global point maps, and depth~\citep{wang2025vggt,cut3r,fast3r}. While point maps are insufficient to ensure consistency, redundant targets can improve pose accuracy but often introduce entanglement that compromises it.
In contrast, our experiments (\cref{tab:ablation_minimal_targets}) show that a depth-ray representation forms a minimal yet sufficient target set for capturing both scene structure and camera motion, outperforming alternatives like point maps or more complex outputs. 
However, recovering camera pose from the ray map at inference is computationally costly. We address this by adding a lightweight camera head, $\camerahead$. This transformer operates on camera tokens to predict the field of view ($\fovparam \in \mathbb{R}^2$), rotation as a quaternion ($\rotquat \in \mathbb{R}^4$), and translation ($\cameratrans \in \mathbb{R}^3$). Since it processes only one token per view, the added cost is negligible.

\subsection{Architecture} \label{sec:archi}
We now detail the architecture of Depth Anything 3, which is illustrated in \cref{fig:pipeline}. The network is composed of three main components: a single transformer model as the backbone, an optional camera encoder for pose conditioning, and a Dual-DPT head for generating predictions.

\textbf{Single transformer backbone.} We use a Vision Transformer with $\transformerlayers$ blocks, pretrained on large-scale monocular image corpora (\textit{e.g.}, DINOv2~\cite{oquab2023dinov2}). 
Cross-view reasoning is enabled without architectural changes via an input-adaptive self-attention, implemented by rearranging input tokens. 
We divide the transformer into two groups of sizes $\singleviewlayers$ and $\globallayers$. 
The first $\singleviewlayers$ layers apply self-attention within each image, while the subsequent $\globallayers$ layers alternate between cross-view and within-view attention, operating on all tokens jointly through tensor reordering. 
In practice, we set $\singleviewlayers:\globallayers = 2:1$ with $\transformerlayers = \singleviewlayers + \globallayers$.
As shown in our ablation study in \cref{tab:abl_dav3}, this configuration provides the optimal trade-off between performance and efficiency compared to other arrangements. 
This design is input-adaptive: with a single image, the model naturally reduces to monocular depth estimation without extra cost.

\textbf{Camera condition injection.} 
To seamlessly handle both posed and unposed inputs, we prepend each view with a camera token $\cameratoken_i$. 
If camera parameters $(\cameraint_i, \camerarot_i, \cameratrans_i)$ are available, the token is obtained via a lightweight MLP $\cameraencoder$: $\cameratoken_i = \cameraencoder(\fovparam_i, \rotquat_i, \cameratrans_i)$. 
Otherwise, a shared learnable token $\learnabletoken$ is used. 
Concatenated with patch tokens, these camera tokens participate in all attention operations, providing either explicit geometric context or a consistent learned placeholder.

\begin{wrapfigure}{r}{0.5\textwidth}
    \centering
    \includegraphics[width=0.48\textwidth]{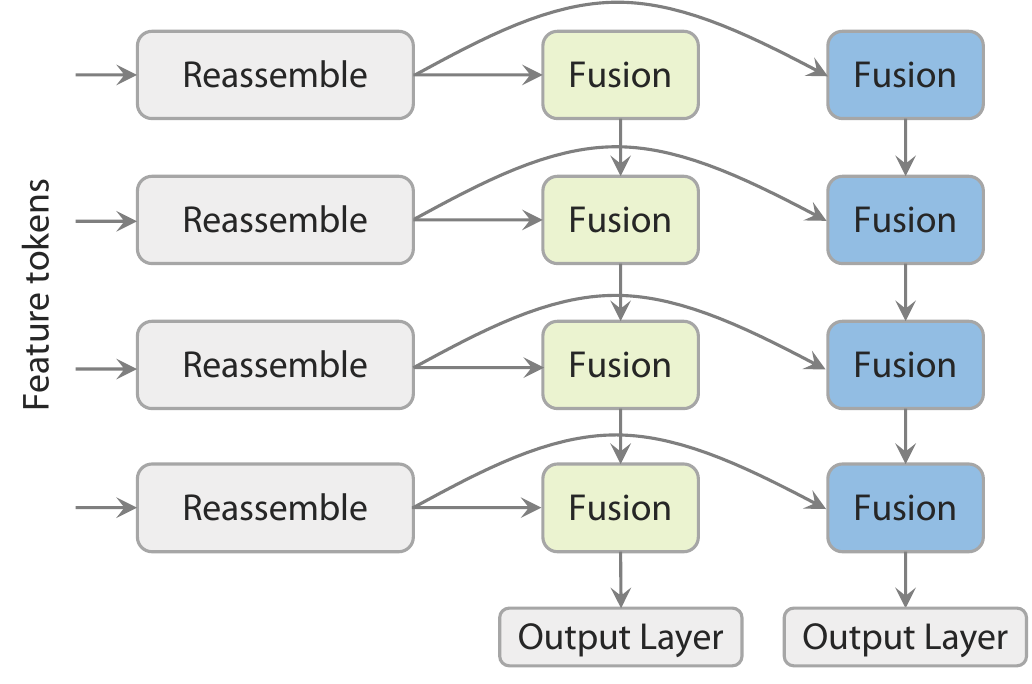}
    \caption{\textbf{Dual-DPT Head.} Two branchs share reassembly modules for better outputs alignment.}
    \label{fig:wrap_example}
\end{wrapfigure}

\textbf{Dual-DPT head.} 
For the final prediction stage, we propose a novel Dual-DPT head that jointly produces dense depth and ray values. As shown in \cref{tab:ablation_minimal_targets}, this design is both powerful and efficient. Given a set of features from the backbone, the Dual-DPT head first processes them through a shared set of reassembly modules. Subsequently, the processed features are fused using two distinct sets of fusion layers: one for the depth branch and one for the ray branch. Finally, two separate output layers produce the final depth and ray map predictions. This architecture ensures that both branches operate on the same set of processed features, differing only in the final fusion stage. Such a design encourages strong interaction between the two prediction tasks, while avoiding redundant intermediate representations.

\subsection{Training} \label{sec:training}

\textbf{Teacher-student learning paradigm.} 
Our training data comes from diverse sources, including real-world depth captures, 3D reconstructions, and synthetic datasets. 
Real-world depth is often noisy and incomplete (\cref{fig:poordataset}), limiting its supervisory value. 
To mitigate this, we train a monocular relative depth estimation ``teacher'' model solely on synthetic data to generate high-quality pseudo-labels. 
These pseudo-depth maps are aligned with the original sparse or noisy ground truth via RANSAC least squares, enhancing label detail and completeness while preserving geometric accuracy. 
We term this model Depth-Anything-3-Teacher, trained on a large synthetic corpus covering indoor, outdoor, object-centric, and diverse in-the-wild scenes to capture fine geometry. 
We detail our teacher design in the~\cref{sec:teacher}.

\newcommand{\errd}{e^{\mathrm{D}}}
\newcommand{\nz}{Z_{\Omega}}

\paragraph{Training objectives.} 
Following the formulation in \cref{sec:formulation}, 
our model $\model$ maps an input $\inputset$ to a set of outputs comprising a depth map $\preddepth$, a ray map $\predray$, and an optional camera pose $\textcolor{gray}{\predcamera}$:
$\model: \inputset \mapsto \{\preddepth, \predray, \textcolor{gray}{\predcamera}\}$.
The gray color indicates that $\textcolor{gray}{\predcamera}$ is an optional output, included primarily for practical convenience. Prior to loss computation, all ground-truth signals are normalized by a common scale factor. This scale is defined as the mean $\ell_{2}$ norm of the valid reprojected point maps $\worldpoint$, a step that ensures consistent magnitude across different modalities and stabilizes the training process. The overall training objective is defined as a weighted sum of several terms:
\begin{equation*}
\totalloss = \depthloss(\preddepth, \depthmap) + \rayloss(\predray, \raymap) + \pointloss(\preddepth \odot \raydir + \rayorigin, \worldpoint) + \textcolor{gray}{\cameraweight \cameraloss (\predcamera, \cameravec)}
+ \gradweight \gradloss(\preddepth, \depthmap),
\label{eq:loss}
\end{equation*}
\begin{equation*}
    \depthloss(\preddepth, \depthmap \,;\, D_c)
    = \tfrac{1}{\nz} \sum_{p \in \Omega} m_p \left( D_{c,p}\, \big|\preddepth_p - \depthmap_p\big| - \lambda_c \log D_{c,p} \right),
\end{equation*}
where $D_{c,p}$ denotes the confidence of depth $D_p$.  All loss terms are based on the $\ell_{1}$ norm, with weights set to $\gradweight=1$ and $\cameraweight=1$. The gradient loss,  $\gradloss$, penalizes the depth gradients:
\begin{equation}
\gradloss(\preddepth, \depthmap) = ||\gradx \preddepth - \gradx \depthmap||_{1} + ||\grady \preddepth - \grady \depthmap||_{1},
\end{equation}
where $\gradx$ and $\grady$ are the horizontal and vertical finite difference operators. This loss preserves sharp edges while ensuring smoothness in planar regions.
In practice, we set $\alpha=1$ and $\beta = 1$.

\subsection{Implementation Details}

\paragraph{Traing datasets.}
We provide our training datasets in Table~\ref{tab:datasets}.
Note that for datasets with potential overlap between training and testing (ScanNet++), we ensure a strict separation at the scene level, i.e., scenes in training and testing are mutually exclusive. Note that using Scannet++ for training is fair to other methods, as it is widely used for training in ~\cite{wang2025vggt,dust3r}.
\begin{table}[ht]
\centering
\caption{Datasets used in \longname, including number of scenes, data type.}
\label{tab:datasets}
\begin{tabular}{l|llc}
\toprule
\textbf{Usage} & \textbf{Dataset} & \textbf{\#Scenes} & \textbf{Data Type} \\
\midrule
\multirow{5}{*}{\makecell{Pose-geometry \\ benchmark}} 
  & HiRoom (ours)        & 29  & Synthetic \\
  & ETH3D~\citep{schops2017eth3d}     & 11  & LiDAR \\
  & DTU~\citep{aanaes2016dtu}         & 22  & LiDAR \\
  & 7Scenes~\citep{shotton2013sevenscene} & 7   & LiDAR \\
  & ScanNet++~\citep{yeshwanth2023scannet++} & 20  & LiDAR \\
\midrule
\multirow{21}{*}{\makecell{Pose-geometry \\ Training}}
  & AriaDigitalTwin~\citep{pan2023aria} & 237   & Synthetic \\
  & AriaSyntheticENV~\citep{pan2023aria} & 99950  &  Synthetic \\
  & ArkitScenes~\citep{baruch2021arkitscenes}      & 4388 & LiDAR \\
  & BlendedMVS~\citep{yao2020blendedmvs} & 503 & 3D Recon \\
  & Co3dv2~\citep{reizenstein2021common} & 30616 & Colmap\\
  & DL3DV~\citep{ling2024dl3dv} & 6379 & Colmap \\
  & HyperSim~\citep{roberts2021hypersim} & 344 & Synthetic \\
  & MapFree~\citep{arnold2022map} & 921 & Colmap \\
  & MegaDepth~\citep{li2018megadepth} & 268 & Colmap \\ 
  & MegaSynth~\citep{jiang2025megasynth} & 6049 & Synthetic \\ 
  & MvsSynth~\citep{DeepMVS} & 121 & Synthetic \\
  & Objaverse~\citep{deitke2023objaverse} & 505557 & Synthetic \\ 
  & Omniobject~\citep{wu2023omniobject3d} & 5885 & Synthetic\\ 
  & OmniWorld~\citep{zhou2025omniworld} & 1039 & Synthetic\\ 
  & PointOdyssey~\citep{zheng2023pointodyssey} & 44 & Synthetic \\ 
  & ReplicaVMAP~\citep{straub2019replica} & 17 & Synthetic \\
  & ScanNet++~\citep{yeshwanth2023scannet++} & 230 & LiDAR \\
  & ScenenetRGBD~\citep{mccormac2017scenenet} & 16866 & Synthetic \\
  & TartanAir~\citep{tartanair2020iros} & 355 & Synthetic \\
  & Trellis~\citep{xiang2024structured} & 557408 & Synthetic \\
  & vKitti2~\citep{cabon2020vkitti2} & 50 & Synthetic \\
  & WildRGBD~\citep{xia2024rgbd} & 23050 & LiDAR \\
\bottomrule
\end{tabular}

\end{table}

\paragraph{Training details.}
We train our model on 128 H100 GPUs for 200k steps, using an 8k-step warm-up and a peak learning rate of $2 \times10^{-4}$. 
The base resolution is $504\times504$, which is divisible by 2, 3, 4, 6, 9 and 14, making it more compatible with common photo aspect ratios such as 2:3, 3:4, and 9:16.
Training image resolutions are randomly sampled from $504\times504$, $504\times378$, $504\times336$, $504\times280$, $336\times504$, $896\times504$, $756\times504$, $672\times504$. For the $504\times504$ resolution, the number of views is sampled uniformly from [2, 18]. The batch size is dynamically adjusted to keep the token count per step approximately constant. Supervision transitions from ground-truth depth to teacher-model labels at 120k steps. Pose conditioning is randomly activated during training with probability 0.2.

\section{Teacher-Student Learning}
\begin{wrapfigure}{r}{0.5\textwidth} 
    \vspace{-5mm}
    \includegraphics[width=0.48\textwidth]{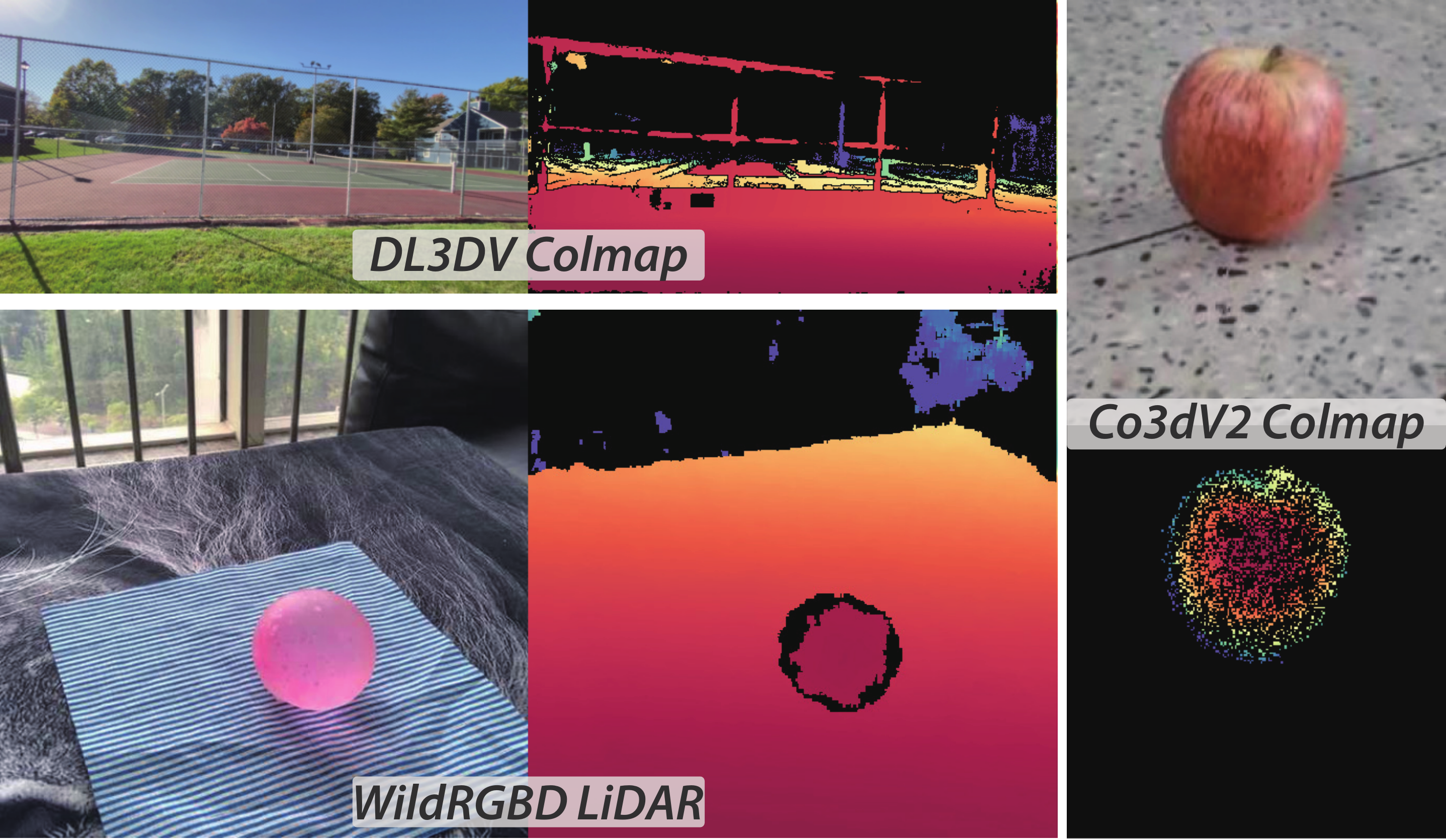}
    \caption{\textbf{Poor quality real-world datasets.} We show some examples of the poor quality real-world datasets. }
    \label{fig:poordataset}
    \vspace{-8mm}
\end{wrapfigure}

As shown in ~\cref{fig:poordataset}, the real-world datasets are of poor quality, thus we train the teacher model exclusively on synthetic data to provide supervision for real-world data.
Our teacher model is trained as a monocular relative depth predictor. During inference or supervision, noisy ground-truth depth can be used to provide scale and shift parameters, allowing for the alignment of the predicted relative depth with absolute depth measurements.

\subsection{Constructing the Teacher Model} \label{sec:teacher}

Building upon Depth Anything 2~\cite{yang2024depth}, we extend the approach in several key aspects, including both data and representation. We observe that expanding the training corpus yields clear improvements in depth estimation performance, substantiating the benefits of data scaling. Furthermore, while our revised depth representation may not show striking improvements on standard 2D evaluation metrics, it leads to qualitatively better 3D point clouds, exhibiting fewer geometric distortions and more realistic scene structures.
It is important to note that our teacher network backbone is directly aligned with the above DA3 framework, consisting solely of a DINOv2 vision transformer with a DPT decoder---no specialized architectural modifications are introduced. We elaborate on the full design and implementation details in the following sections.

\noindent{\bf Data scaling.} We train the teacher model exclusively on synthetic data to achieve finer geometric detail. The synthetic datasets used in DA2 are relatively limited. In DA3, we substantially expand the training corpus to include: Hypersim~\citep{roberts2021hypersim}, TartanAir~\citep{tartanair2020iros}, IRS~\citep{wang2019irs}, vKITTI2~\citep{cabon2020vkitti2}, BlendedMVS~\citep{yao2020blendedmvs}, SPRING~\citep{mehl2023spring}, MVSSynth~\citep{DeepMVS}, UnrealStereo4K~\citep{zhang2018unrealstereo}, GTA-SfM~\citep{wang2020flow}, TauAgent~\citep{gil2021online}, KenBurns~\cite{niklaus20193d}, MatrixCity~\citep{li2023matrixcity}, EDEN~\citep{le2021eden}, ReplicaGSO~\citep{replica19arxiv}, UrbanSyn~\citep{GOMEZ2025130038}, PointOdyssey~\citep{zheng2023pointodyssey}, Structured3D~\citep{zheng2020structured3d}, Objaverse~\citep{deitke2023objaverse}, Trellis~\citep{xiang2024structured}, and OmniObject~\citep{wu2023omniobject3d}. This collection spans indoor, outdoor, object-centric, and diverse in-the-wild scenes, improving generalization of the teacher model.

\noindent{\bf Depth representation.} Unlike DA2, which predicts scale–shift-invariant disparity, our teacher outputs scale–shift-invariant depth. Depth is preferable for downstream tasks, such as metric depth estimation and multiview geometry, that directly operate in depth space rather than disparity.
To address depth's reduced sensitivity for near-camera regions comparing to disparity, we predict exponential depth instead of linear depth, enhancing discrimination at small distances.

\noindent{\bf Training objectives.} 
For geometric supervision, in addition to a standard depth-gradient loss, we adopt ROE alignment with the global–local loss introduced in~\cite{wang2025moge}. To further refine local geometry, we introduce a distance-weighted surface-normal loss. For each center pixel, we sample four neighboring points and compute unnormalized normals $n_i$. We then weight these normals by:
\begin{equation}
  w_i= \sum _ {j=0}^ 4 \parallel n_j\parallel - \parallel n_i \parallel,
  \label{eq:normal_weight}
\end{equation}
which downweights contributions from neighbors farther from the center, yielding a mean normal closer to the true local surface normal:
\begin{equation}
 n_m= \sum _ {i=0}^ {4} w_i \frac {n_i}{\parallel n_i \parallel},
 \label{eq:normal_mean}
\end{equation}
The final normal loss is
\begin{equation}
 \mathcal{L}_{N} = \mathcal{E}(\hat{n}_{m}, n_m) + \sum _ {i=0}^{4} \mathcal{E}(\hat{n}_i, n_i)
 \label{eq:normal_loss}
\end{equation}
where $\mathcal{E}$ denotes the angular error between normals. Ground truth is undefined in sky regions and in background areas of object-only datasets. To prevent these regions from degrading the depth prediction and to facilitate downstream use, we jointly predict a sky mask and an object mask aligned with the depth output, supervised with MSE loss. The overall training objective is
\begin{equation}
 \mathcal{L}_{T} = \gradweight \mathcal{L}_\text{grad} + \mathcal{L}_\text{gl} + \mathcal{L}_{N} + \mathcal{L}_\text{sky} + \mathcal{L}_\text{obj}
 \label{eq:teacher_loss}
\end{equation}
where $\gradweight = 0.5$. Here, $\mathcal{L}_\text{grad}$, $\mathcal{L}_\text{gl}$, $\mathcal{L}_\text{sky}$, and $\mathcal{L}_\text{obj}$ denote the gradient loss, global–local loss, sky-mask loss, and object-mask loss, respectively.

\subsection{Teaching Depth Anything 3}
Real-world datasets are crucial for generalizing camera pose estimation, yet they rarely provide clean depths; supervision is often noisy or sparse (\cref{fig:poordataset}). Depth Anything 3 Teacher provides high-quality relative depth, which we align to noisy metric measurements (e.g., COLMAP or active sensors) via a robust ransac scale–shift procedure. Let \(\tilde{\depthmap}\) denote the teacher's relative depth and \(\depthmap\) the available sparse depth with validity mask \(m_p\) over domain \(\Omega\). We estimate scale \(s\) and shift \(t\) by RANSAC least squares, using an inlier threshold equal to the mean absolute deviation from the residual median:
\begin{equation}
    (\hat{s},\hat{t}) 
    = \arg\min_{s>0,\,t} \sum_{p\in\Omega} m_p\,\big( s\,\tilde{\depthmap}_p + t - \depthmap_p\big)^2,\quad
    \depthmap^{T\rightarrow M} = \hat{s}\,\tilde{\depthmap} + \hat{t}.\label{eq:ss_align}
\end{equation}
The aligned \(\depthmap^{T\rightarrow M}\) provides scale-consistent and pose–depth coherent supervision for Depth Anything 3, complementing our joint depth–ray objectives and improving real-world generalization, as evidenced in \cref{fig:vis_teacher_sup}.

\subsection{Teaching Monocular Model}

We additionally train a monocular depth model under a teacher–student paradigm.
We follow the DA2 framework, training the monocular student on unlabeled images with teacher-generated pseudo-labels. The key difference from DA2 lies in the prediction target: our student predicts depth maps, whereas DA2 predicts disparity. We further supervise the student with the same loss used for the teacher, applied to the pseudo-depth labels.
The monocular model also predicts relative depth. Trained solely on unlabeled data with teacher supervision, it achieves state-of-the-art performance on standard monocular depth benchmarks as shown in \cref{tab:student_mono}.

\subsection{Teaching Metric Model}
Next, we demonstrate that our teacher model can be used for training a metric depth estimation model with sharp boundaries. Following Metric3Dv2~\cite{hu2024metric3dv2}, we apply canonical camera space transformation to address depth ambiguity caused by varying focal lengths. Specifically, we rescale the ground-truth depth using ratio $f^c/f$, where $f^c$ and $f$ denote the canonical focal length and camera focal length, respectively. To ensure sharp details, we employ Teacher model's prediction as training labels. We align the scale and shift of the teacher model's predicted depths with the ground-truth metric depth labels for supervision. 

\paragraph{Training dataset.}
We trained our metric depth model on 14 datasets, including Taskonomy~\cite{zamir2018taskonomy}, DIML (Outdoor)~\cite{cho2021diml}, DDAD~\cite{ddad}, Argoverse~\cite{wilson2023argoverse}, Lyft~[], PandaSet~\cite{xiao2021pandaset}, Waymo~\cite{Sun_2020_CVPR}, ScanNet++~\cite{yeshwanth2023scannet++}, ARKitScenes~\cite{baruch2021arkitscenes}, Map-free~\cite{arnold2022map}, DSEC~\cite{Gehrig21ral}, Driving Stereo~\cite{yang2019drivingstereo} and Cityscapes~\cite{cordts2016cityscapes} datasets. For stereo datasets, we leverage the prediction of FoundationStereo~\cite{wen2025foundationstereo} as training labels. 

\paragraph{Implementation Details.}
The training largely follows that of the monocular teacher model. All images are trained at a base resolution of 504 with varying aspect ratios (1:1, 1:2, 16:9, 9:16, 3:4, 1:1.5, 1.5:1, 1:1.8). We employ AdamW optimizer and set the learning rate for encoder and decoder to 5e-6 and 5e-5, respectively. We apply random rotation augmentation where training images are rotated at 90 or 270 degree with 5\% probability. We set canonical focal length $f^c$ to 300. We use the aligned prediction from teacher model as supervision. With a probability of 20\%, we use the original ground-truth labels for training. We train with batch size of 64 for 160K iterations. The training objective is a weighted sum of depth loss $\mathcal{L}_{depth}$, $\mathcal{L}_{grad}$ and sky-mask loss $\mathcal{L}_{sky}$.

\section{Application: Feed-Forward 3D Gaussian Splattings}
\subsection{Pose-Conditioned Feed-Forward 3DGS}
Inspired by human spatial intelligence, we believe that consistent depth estimation can greatly enhance downstream 3D vision tasks. We choose feed-forward novel view synthesis (FF-NVS) as the demonstration task, given its growing attention driven by advances in neural 3D representations (\ie, we choose 3DGS) and its relevance to numerous applications. Adhere to the minimal modeling strategy, we perform FF-NVS by \textit{fine-tuning} with an added DPT head (GS-DPT) to infer pixel-aligned 3D Gaussians~\cite{charatan2024pixelsplat,chen2024mvsplat}.

\textbf{GS-DPT head.} 
Given visual tokens for each view extracted via our single transformer backbone (\cref{sec:archi}), GS-DPT predicts the camera-space 3D Gaussian parameters $\{ \sigma_i, \mathbf{q}_i, \mathbf{s}_i, \mathbf{c}_i \}_{i=1}^{H \times W} $, where $\sigma_i$, $\mathbf{q}_i \in \mathbb{H}$, $\mathbf{s}_i \in \mathbb{R}^3$, $\mathbf{c}_i \in \mathbb{R}^3$ denote the opacity, rotation quaternion, scale, and RGB color of the $i$-th 3D Gaussian, respectively. Among them, $\sigma_i$ is predicted by the confidence head, while others are predicted by the main GS-DPT head. The estimated depth is unprojected to world coordinates to obtain the global positions ${\worldpoint}_i \in \mathbb{R}^3 $ of the 3D Gaussians. These primitives are then rasterized to synthesize novel views from given camera poses. 

\textbf{Training objectives.} 
The NVS model is fine-tuned with two training objectives, namely photometric loss (\ie, $\mseloss$ and $\lpipsloss$) on rendered novel views and scale-shift-invariant depth loss $\depthloss$ on the estimated depth of observed views, following the teacher–student learning paradigm (\cref{sec:training}).

\subsection{Pose-Adaptive Feed-Forward 3DGS}

Unlike the above pose-conditioned version intended to benchmark DA3 as a strong feed-forward 3DGS backbone, we also present an alternative better suited to in-the-wild evaluation. This version is designed to integrate seamlessly with DA3 using \textit{identical} pretrained weights, enabling novel view synthesis with or without camera poses, and across varying resolutions and input view counts.

\textbf{Pose-adaptive formulation.} Rather than assuming that all input images are uncalibrated~\cite{smart2024splatt3r,ye2024no,flare,jiang2025anysplat}, we adopt a \textit{pose-adaptive} design that accepts both posed and unposed inputs, yielding a flexible framework that works \textit{with or without poses}. Two design choices are required to achieve this: 1) all 3DGS parameters are predicted in local camera space. 2) the backbone must handle posed and unposed images seamlessly. Our DA3 backbone satisfies both requirements (\cref{sec:archi}). In particular, when poses are available, we scale (via \cite{umeyama2002least}) and unproject the predicted depth and camera-space 3DGS to world space to align with them. When poses are not available, we directly use the predicted poses for the unprojection to world space.

To reduce the trade-off between accurate surface geometry and rendering quality~\cite{guedon2024sugar}, we predict an additional depth offset in the GS-DPT head. For more in-the-wild robustness, we replace per 3D Gaussian color with spherical harmonic coefficients to reduce conflicts with geometry via modeling view-dependent surface.

\textbf{Enhanced training strategies.} To avoid unstable training, we initialize the DA3 backbone from pretrained weights and \textit{freeze} it when training, tuning only the GS-DPT head. To improve in-the-wild performance, we train with varying image resolutions and varying numbers of context views. Specifically, higher-resolution inputs are paired with fewer context views and lower-resolution inputs with more views, which stabilizes training while supporting diverse evaluation scenarios.

\subsection{Implementation Details}

\paragraph{Training datasets.} 
For training the NVS model, we leverage the large-scale DL3DV dataset~\citep{ling2024dl3dv}, which provides diverse real-world scenes with camera poses estimated by COLMAP. We use 10,015 scenes from DL3DV for training the feed-forward 3DGS model. To ensure fair evaluation, we strictly maintain exclusivity between training and testing splits: the 140 DL3DV scenes used for benchmarking are completely disjoint from the training set, preventing any data leakage.

\section{Visual Geometry Benchmark}

We further introduce a visual geometry benchmark to assess geometry prediction models. It directly evaluates pose accuracy, 
depth via reconstruction accuracy and visual rendering quality. 

\subsection{Benchmark Pipeline}

\paragraph{Pose estimation.} 
For each scene, we select all available images; if the total number exceeds the limit, we randomly sample 100 images using a fixed random seed. The selected images are then processed through a feed-forward model to generate consistent pose and depth estimations, after which the pose accuracy is computed.

\paragraph{Geometry estimation.} For the same image set, we perform reconstruction using the predicted poses together with the predicted depths. To align the reconstructed point cloud with the ground-truth, we employ evo~\citep{umeyama2002least} to align the predicted poses to the ground-truth poses, obtaining a transformation that maps the reconstruction into the ground-truth coordinate system. To improve robustness, we adopt a RANSAC-based alignment procedure. Specifically, we repeatedly apply evo on randomly sampled pose subsets and evaluate each candidate transformation by counting the number of inlier poses, where inliers are defined as those with translation errors below the median of the overall pose deviations. The transformation with the largest inlier set is then chosen and applied to fuse the aligned predicted point cloud with the predicted depth maps by TSDF fusion. 
Finally, reconstruction quality is assessed by comparing the aligned reconstruction with the ground-truth point cloud using the metrics described in Sec.~\ref{subsection:appendix:reconstruction_metrics}.

\paragraph{Visual rendering.}
For each testing scene, the number of images typically ranges from 300 to 400 across all benchmark datasets. We sample one out of every 8 images as target novel views for evaluation. From the remaining viewpoints, we use COLMAP camera poses provided by each dataset and apply farthest point sampling, considering both camera translation and rotation distances, to select 12 images as input context views. For DL3DV, we use the official Benchmark set for testing. For Tanks and Temples, all Training Data scenes are included except \texttt{Courthouse}. For MegaDepth, we select scenes numbered from \texttt{5000} to \texttt{5018}, as these are most suitable for NVS.

\subsection{Metrics}

\paragraph{Pose metrics.} For assessing pose estimation, we follow the evaluation protocol introduced in~\cite{wang2025vggt,wang2023posediffusion} and report results using the AUC. This metric is derived from two components: Relative Rotation Accuracy (RRA) and Relative Translation Accuracy (RTA). RRA and RTA quantify the angular deviation in rotation and translation, respectively, between two images. Each error is compared against a set of thresholds to obtain accuracy values. AUC is then computed as the integral of the accuracy–threshold curve, where the curve is determined by the smaller of RRA and RTA at each threshold. To illustrate performance under different tolerance levels, we primarily report results at thresholds of 3 and 30.

\paragraph{Reconstrution metrics.}
\label{subsection:appendix:reconstruction_metrics}
Let $\mathcal{G}$ denote the ground-truth point set and $\mathcal{R}$ the reconstructed point set under evaluation. We measure accuracy using $\text{dist}(\mathcal{R}\rightarrow \mathcal{G})$ and completeness using $\text{dist}(\mathcal{G}\rightarrow \mathcal{R})$ following~\cite{aanaes2016dtu}. The Chamfer Distance (CD) is then defined as the average of these two terms. Based on these distances, we define the precision and recall of the reconstruction $\mathcal{R}$ with respect to a distance threshold $d$. Precision is given by $\frac{1}{|\mathcal{R}|}\sum \big[\text{dist}(\mathcal{R}_i \rightarrow \mathcal{G}) < d \big]$, and recall by $\frac{1}{|\mathcal{G}|}\sum \big[\text{dist}(\mathcal{G}_i \rightarrow \mathcal{R}) < d \big]$, where $\left[\cdot \right]$ denotes the Iverson bracket~\cite{knapitsch2017tnt}. To jointly capture both measures, we report the F1-score, computed as $\text{F1} = \frac{2\times \text{precision}\times \text{recall}}{\text{precision}+\text{recall}}$.

\subsection{Datasets}

Our benchmark is built on five datasets: HiRoom~\citep{SVLightVerse2025}, ETH3D~\citep{schops2017eth3d}, DTU~\citep{aanaes2016dtu}, 7Scenes~\citep{shotton2013sevenscene}, and ScanNet++~\citep{yeshwanth2023scannet++}. Together, they cover diverse scenarios ranging from object-centric captures to complex indoor and outdoor environments, and are widely adopted in prior research. Below, we present more details about the dataset preparation process. 

\textbf{HiRoom} is a Blender-rendered synthetic dataset comprising 30 indoor living scenes created by professional artists.
We use a threshold $d$ of 0.05m for the F1 reconstruction metric calculation. 
For TSDF fusion, we set the parameters voxel size to 0.007m.

\textbf{ETH3D} provides high-resolution indoor and outdoor images with ground-truth depth from laser sensors. We aggregate the ground-truth depth maps with TSDF fusion for GT 3D shapes. We select 11 scenes: \texttt{courtyard}, \texttt{electro}, \texttt{kicker}, \texttt{pipes}, \texttt{relief}, \texttt{delivery area}, \texttt{facade}, \texttt{office}, \texttt{playground}, \texttt{relief 2}, \texttt{terrains}, for the benchmark. All frames are used in the evaluation. 
We use a threshold $d$ of 0.25 for the F1 reconstruction metric calculation. 
For TSDF fusion, we set the parameters voxel size to 0.039m.

\textbf{DTU} is an indoor dataset consisting of 124 different objects, each scene is recorded from 49 views.  It provides ground-truth point clouds collected under well-controlled conditions. We evaluate models on the 22 evaluation scans of the DTU dataset following~\cite{yao2018mvsnet}. We adopt the RMBG 2.0~\cite{BiRefNet} to remove meaningless background pixels and use the default depth fusion strategy proposed in \cite{zhang2023geomvsnet}. All frames are used in the evaluation. 

\textbf{7Scenes} is a challenging real-world dataset, consisting of low-resolution images with severe motion blurs for in-door scenes. We follow the implementation in~\cite{zhu2024nicerslam} to fuse RGBD images with TSDF fusion and prepare ground-truth 3D shapes. 
We downsample the number of frames for each scene by 11 to faciliate evaluation.
We use a threshold $d$ of 0.05m for the F1 reconstruction metric calculation. 
For TSDF fusion, we set the parameters voxel size to 0.007m.

\textbf{ScanNet++} is an extensive indoor dataset providing high-resolution images, depth maps from iPhone LiDAR, and high-resolution depth maps sampled from reconstructions of laser scans. We select 20 scenes for the benchmark. As depth maps from iPhone LiDAR lack of invalid ground-truth indicators, we use depth maps sampled from reconstructions of laser scans as ground-truth depth by default. We aggregate the ground-truth depth maps with TSDF fusion for GT 3D shapes. We downsample the number of frames for each scene by 5 to faciliate evaluation. We use a threshold $d$ of 0.05m for the F1 reconstruction metric calculation. 
For TSDF fusion, we set the parameters voxel size to 0.02m.

\paragraph{Visual rendering quality.} \label{sec:bm_ffnvs}
We evaluate visual rendering quality on diverse large-scale scenes. 
We introduce a new NVS benchmark built from three datasets, including DL3DV~\citep{ling2024dl3dv} with 140 scenes, Tanks and Temples~\citep{knapitsch2017tanks} with 6, and MegaDepth~\citep{li2018megadepth} with 19, each spanning around 300 sampled frames. Ground truth camera poses, estimated with COLMAP, are used directly to ensure accurate and fair comparison across diverse models. 
We report PSNR, SSIM, and LPIPS metrics on rendered novel views using given camera poses.

\newpage
\section{Experiments}

\begin{table}[t]
  \caption{
      \textbf{Comparisons with SOTA methods on pose accuracy.} We report both Auc3 $\uparrow$ and Auc30 $\uparrow$ metrics. The top-3 results are highlighted as \textfirst{first}, \secondtext{second}, and \thirdtext{third}.
}
  \label{tab:comp_pose}
  \begin{center}
  \resizebox{\linewidth}{!}{
  \begin{tabular}{lccccccccccc}
      \toprule
      \multirow{2}{*}[-4pt]{\textbf{Methods}} 
      & \multirow{2}{*}[-4pt]{\textbf{Params}} 
      & \multicolumn{2}{c}{\textbf{ \picodataset }} 
      & \multicolumn{2}{c}{\textbf{ ETH3D }} 
      & \multicolumn{2}{c}{\textbf{ DTU }} 
      & \multicolumn{2}{c}{\textbf{ 7Scenes }} 
      & \multicolumn{2}{c}{\textbf{ ScanNet++ }} 
      \\
      \cmidrule(lr){3-4} \cmidrule(lr){5-6} \cmidrule(lr){7-8} \cmidrule(lr){9-10} \cmidrule(lr){11-12} 
      & 
      & Auc3 & Auc30
      & Auc3 & Auc30
      & Auc3 & Auc30
      & Auc3 & Auc30
      & Auc3 & Auc30
        \\
      \midrule
      DUSt3R      & 0.57B & 17.6 & 54.3 & 4.30 & 27.3 & 4.00 & 74.3 & 6.90 & 61.6 & 8.10 & 33.9 \\
      Fast3R      & 0.65B & 25.9 & 77.0 & 8.10 & 44.4 & 9.50 & 79.1 & 19.0 & 78.6 & 17.9 & 72.5 \\
      MapAnything & 0.56B & 17.9 & 82.8 & 19.2 & 77.4 & 6.50 & 72.7 & 12.6 & 79.7 & 20.2 & 84.1 \\ 
      Pi3         & 0.96B & \cellsecond{67.0} & \cellsecond{94.8} & \cellsecond{35.2} & \cellsecond{87.3} &62.5 & 94.9 & \cellthird{25.5} & \cellthird{86.3} & 50.7 & 92.1 \\ 
      VGGT        & 1.19B & 49.1 & 88.0 & 26.3 & 80.8 & \cellsecond{79.2} & \cellfirst{99.8} & 23.9 & 85.0 & \cellsecond{62.6} & \cellsecond{95.1} \\ 
      \midrule
      DA3-Giant   & 1.10B & \cellfirst{80.3} & \cellfirst{95.9} & \cellfirst{48.4} & \cellfirst{91.2} & \cellfirst{94.1} & \cellthird{99.4} & \cellsecond{28.5} & \cellfirst{86.8} & \cellfirst{85.0} & \cellfirst{98.1} \\ 
      DA3-Large   & 0.36B & \cellthird{58.7} & \cellthird{94.2} & \cellthird{32.2} & \cellthird{86.9} & \cellthird{70.2} & \cellsecond{96.7} & \cellfirst{29.2} & \cellsecond{86.6} & \cellthird{60.2} & \cellthird{94.7} \\ 
      DA3-Base    & 0.11B & 19.0 & 83.2 & 15.1 & 74.6 & 60.1 & 95.9 & 20.1 & 82.9 & 25.1 & 83.4 \\ 
      DA3-Small   & 0.03B & 9.49 & 75.2 & 8.59 & 62.1 & 30.6 & 91.2 & 14.0 & 78.7 & 10.9 & 71.9 \\ 
      \bottomrule
  \end{tabular}
  }
  \end{center}
\end{table}

\begin{figure}
    \begin{center}
        \includegraphics[width=\textwidth]{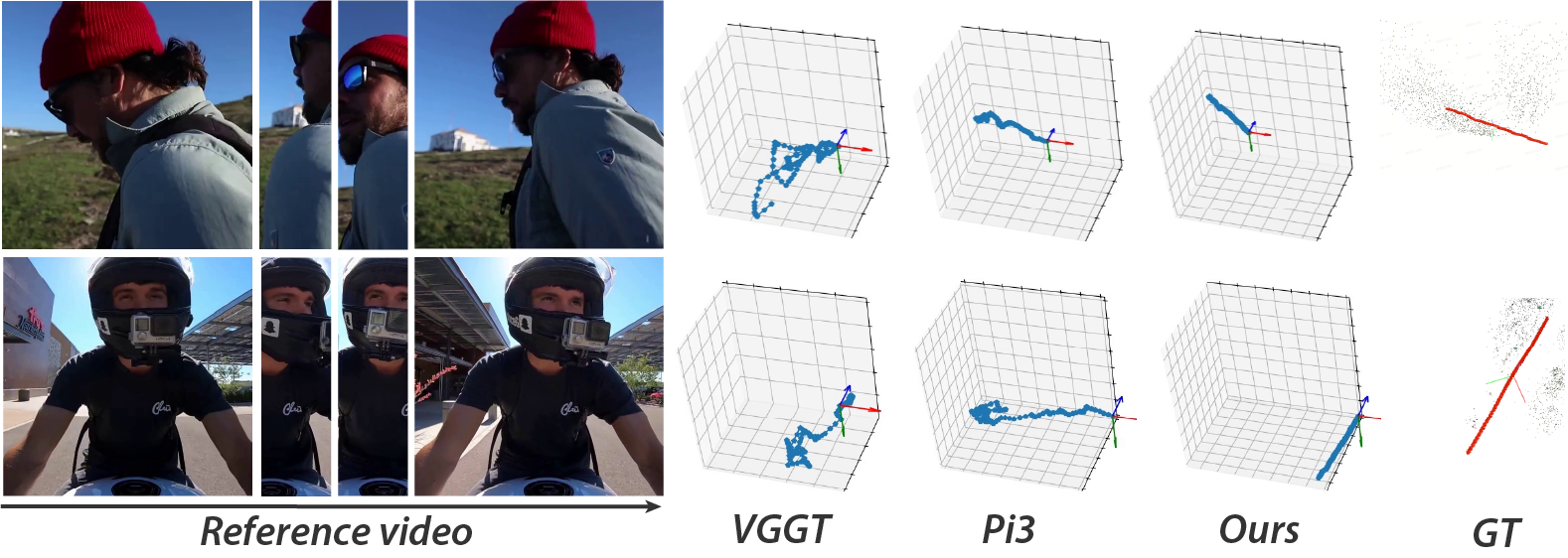}
        \caption{\textbf{Comparisons of pose estimation quality.} Camera trajectories for two videos are shown. Ground-truth trajectories are derived using COLMAP on images with dynamic objects masked. 
        }
        \label{fig:vis_traj}
    \end{center}
\end{figure}

\subsection{Comparison with State of the Art}

\paragraph{Baselines.}
VGGT~\citep{wang2025vggt} is an end-to-end transformer that jointly predicts camera parameters, depth, and 3D points from one or many views. Pi3~\citep{wang2025pi3} further adopts a permutation-equivariant design to recover affine-invariant cameras and scale-invariant point maps from unordered images. MapAnything~\citep{keetha2025mapanything} provides a feed-forward framework that can also take camera pose as input for dense geometric prediction. Fast3R~\citep{yang2025fast3r} extends point-map regression to hundreds or even thousands of images in a single forward pass. Finally, DUSt3R~\citep{wang2024dust3r} tackles uncalibrated image pairs by regressing point maps and aligning them globally. Our method is similar to VGGT~\citep{wang2025vggt}, but adopts a new architecture and a different camera representation, and it is orthogonal to Pi3~\citep{wang2025pi3}.

\paragraph{Pose estimation.} 
As shown in \cref{tab:comp_pose} and \cref{fig:vis_traj}, comparing against five baselines~\citep{dust3r,wang2025vggt,fast3r,wang2025pi3,keetha2025mapanything}, our DA3-Giant model attains the best performance on nearly all metrics, with the only exception being Auc30 on the DTU dataset. Notably, on Auc3 our model delivers at least an $8\%$ relative improvement over all competing methods, and on ScanNet++ it achieves a $33\%$ relative gain over the second-best model.

\begin{table}[t]
    \caption{
        \textbf{Comparisons with SOTA methods on reconstruction accuracy.} 
        For all datasets except DTU, we report the F-Score (\textbf{F1} $\uparrow$). 
        For DTU, we report the chamfer distance (\textbf{CD} $\downarrow$, unit: mm). 
w/o p. and w/ p. denote without pose and with pose, indicating whether ground-truth camera poses are provided for reconstruction.  The top-3 results are highlighted as \textfirst{first}, \secondtext{second}, and \thirdtext{third}.
  }
    \label{tab:comp_recon}
    \begin{center}
    \resizebox{\linewidth}{!}{
    \begin{tabular}{lccccccccccc}
        \toprule
        \multirow{2}{*}[-4pt]{\textbf{Methods}} 
        & \multirow{2}{*}[-4pt]{\textbf{Params}} 
        & \multicolumn{2}{c}{\textbf{ \picodataset }} 
        & \multicolumn{2}{c}{\textbf{ ETH3D }} 
        & \multicolumn{2}{c}{\textbf{ DTU }} 
        & \multicolumn{2}{c}{\textbf{ 7Scenes }} 
        & \multicolumn{2}{c}{\textbf{ ScanNet++ }} 
        \\
        \cmidrule(lr){3-4} \cmidrule(lr){5-6} \cmidrule(lr){7-8} \cmidrule(lr){9-10} \cmidrule(lr){11-12} 
        & 
        & w/o p. & w/ p.
        & w/o p. & w/ p.
        & w/o p. & w/ p.
        & w/o p. & w/ p.
        & w/o p. & w/ p.
          \\
        \midrule
        DUSt3R      & 0.57B & 30.1 & 39.5 & 19.7 & 18.8 & 7.60 & 7.97 & 26.6 & 39.8 & 18.9 & 27.3 \\
        Fast3R      & 0.65B & 40.7 & 48.2 & 38.5 & 50.3 & 6.88 & 8.20 & 41.0 & 49.8 & 37.1 & 53.7 \\
        MapAnything & 0.56B & 32.4 & 69.2 & 54.8 & 71.9 & 7.91 & 3.97 & 44.8 & \cellthird{55.2} & 39.4 & 71.3 \\ 
        Pi3         & 0.96B & \cellsecond{75.8} & \cellthird{85.0} & \cellsecond{72.7} & \cellsecond{80.6} & 3.28 & \cellthird{1.72} & 44.2 & \cellfirst{57.5} & 63.1 & \cellthird{73.3} \\ 
        VGGT        & 1.19B & 56.7 & 70.2 & 57.2 & 66.7 & \cellsecond{2.05} & \cellsecond{1.44} & 47.9 & 51.4 & \cellthird{66.4} & 70.7 \\ 
        \midrule
        DA3-Giant   & 1.10B & \cellfirst{85.1} & \cellfirst{95.6} & \cellfirst{79.0} & \cellfirst{87.1} & \cellfirst{1.85} & 1.85 & \cellsecond{53.5} & \cellsecond{56.5} & \cellfirst{77.0} & \cellfirst{79.3} \\ 
        DA3-Large   & 0.36B & \cellthird{69.5} & \cellsecond{87.1} & \cellthird{65.8} & \cellthird{75.2} & \cellthird{2.08} & \cellfirst{1.23} & \cellfirst{56.3} & 49.2 & \cellsecond{67.9} & \cellsecond{75.7}  \\ 
        DA3-Base    & 0.11B & 25.9 & 71.4 & 49.5 & 66.7 & 2.87 & 2.36 & \cellthird{49.9} & 50.6 & 47.2 & 67.8 \\ 
        DA3-Small   & 0.03B & 18.3 & 52.2 & 41.6 &  63.4 & 5.83 & 2.49 &  41.0 & 46.8 & 32.3 & 53.8 \\ 
        \bottomrule
    \end{tabular}
    }
    \end{center}
\end{table}

\begin{figure}[t]
    \begin{center}
        \includegraphics[width=1\textwidth]{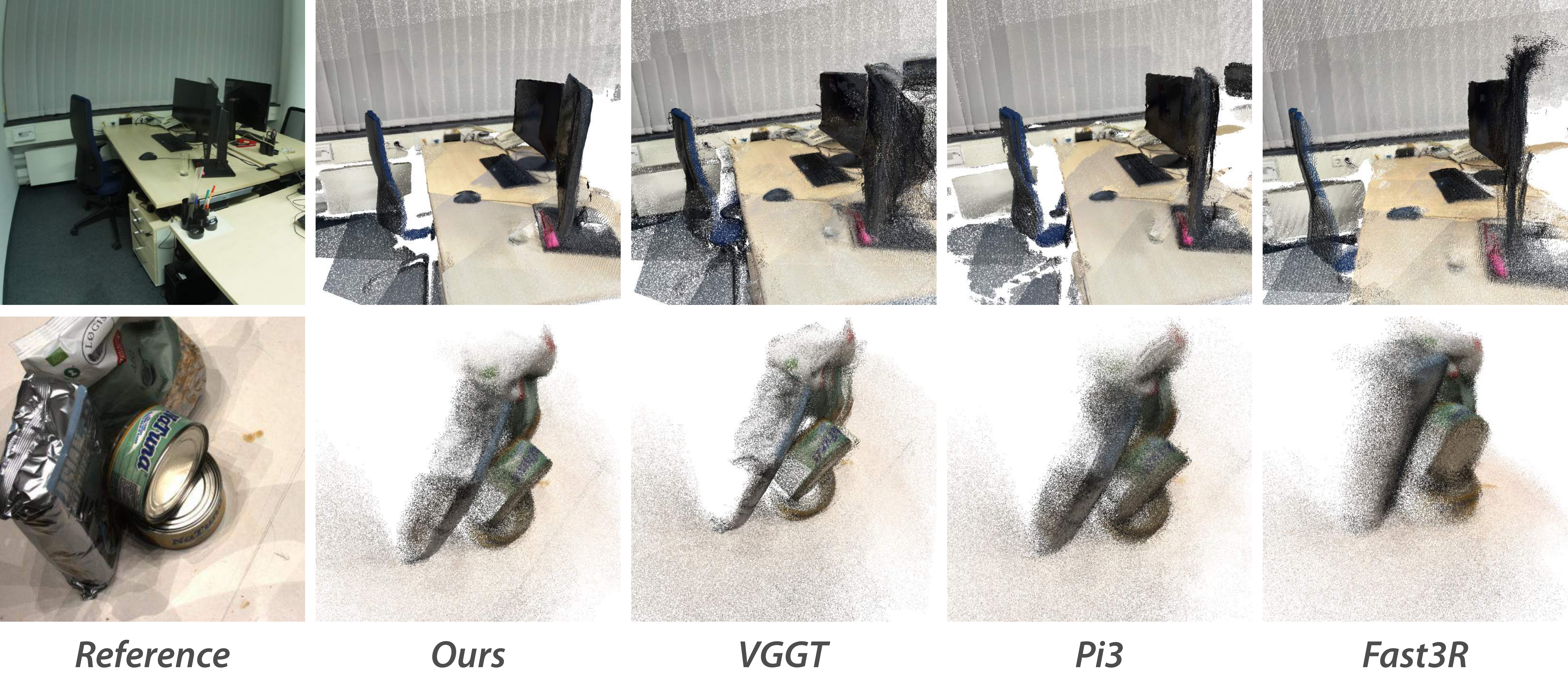}
        \caption{\textbf{Comparisons of point cloud quality.} Our model produces point clouds that are more geometrically regular and substantially less noisy than those generated by other methods.
        }
        \label{fig:pcd_vis}
    \end{center}
\end{figure}

\paragraph{Geometry estimation.}
As shown in \cref{tab:comp_recon},
our {\giantmodel establishes a new SOTA in nearly all scenarios, outperforming all competitors in all five pose-free settings. On average, \giantmodel achieves a relative improvement of \textbf{25.1\%} over VGGT and \textbf{21.5\%} over Pi3. \cref{fig:vis_depth} and \cref{fig:pcd_vis} visualize our predicted depth and recovered point clouds. The results are not only clean, accurate, and complete, but also preserve fine-grained geometric details, clearly demonstrating a superiority over other methods.
Even more notably, our much smaller \largemodel (0.30B parameters) demonstrates remarkable efficiency. 
Despite being 3$\times$ smaller, it surpasses the prior SOTA VGGT (1.19B parameters) in five out of the ten settings, with particularly strong performance on the ETH3D.

When camera poses are available, both our method and MapAnything can exploit them for improved results, and other methods also benefit from ground-truth pose fusion. Our model shows clear gains on most datasets except 7Scenes, where the limited video setting already saturates performance and reduces the benefit of pose conditioning. Notably, with pose conditioning, performance gains from scaling model size are smaller than in pose-free models, \textbf{indicating that pose estimation scales more strongly than depth estimation and requires larger models to fully realize improvements.}

Monocular depth accuracy also reflects geometry quality. As shown in \cref{tab:monodepth}, on the standard monocular depth benchmarks reported in \cite{yang2024depthv2}, our model outperforms VGGT and Depth Anything 2. For reference, we also include the results of our teacher model.

\begin{table}[h]
    \centering
    \caption{\textbf{Monocular depth comparisons.} $\delta_1 \uparrow$}
    \begin{tabular}{lcccccc}
        \toprule
        Method & KITTI & NYU & SINTEL & ETH3D & DIODE & Rank\\
        \midrule
        DA2 & 94.6 & 97.9 & 77.2 & 86.5 & 95.2 & 2.60\\
        VGGT & 91.7 & 97.9 & 67.9 & 97.5 & 95.3 & 3.75    \\
        \shortname& 95.3 & 97.4 & 75.5 & 98.6 & 95.4 & 2.20 \\
        Teacher & 97.2 & 97.9 & 81.4 & 99.8 & 96.6 & 1.00 \\
        \bottomrule
    \end{tabular}
    \label{tab:monodepth}
\end{table}
\begin{figure}[ht]
    \begin{center}
        \includegraphics[width=1\textwidth]{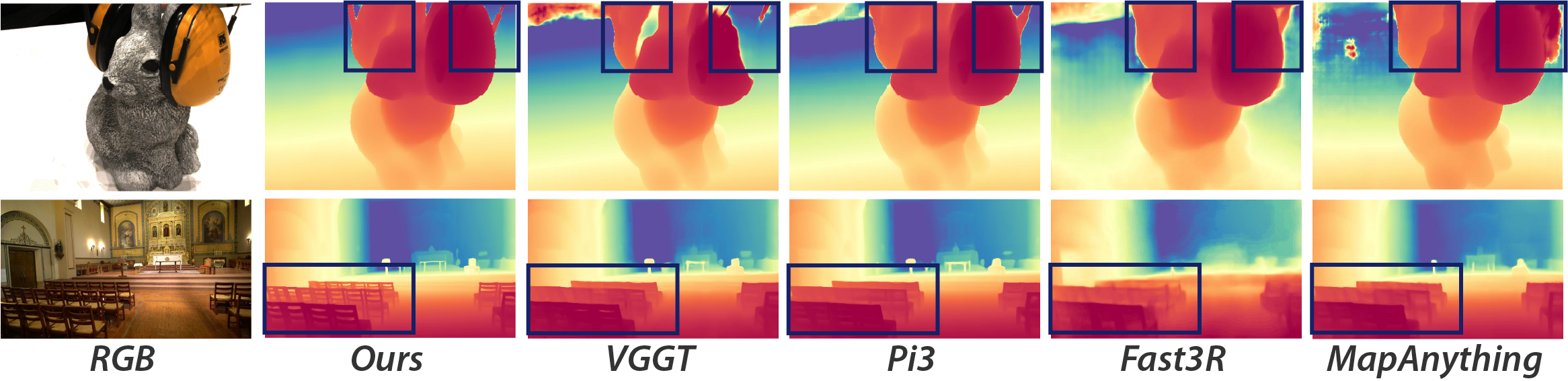}
        \caption{\textbf{Comparisons of depth quality.} Compared with other methods, our depth maps exhibit finer structural detail and higher semantic correctness across diverse scenes.
        }
        \label{fig:vis_depth}
    \end{center}
\end{figure}

\paragraph{Visual rendering.}

To fairly evaluate feed-forward novel view synthesis (FF-NVS), we compare against three recent 3DGS models—pixelSplat~\citep{charatan2024pixelsplat}, MVSplat~\citep{chen2024mvsplat}, and DepthSplat~\citep{xu2025depthsplat}—and further test alternative frameworks by replacing our geometry backbone with Fast3R~\citep{yang2025fast3r}, MV-DUSt3R~\citep{Mv-dust3r+}, and VGGT~\citep{wang2025vggt}. All models are trained on DL3DV-10K training set under a unified protocol and evaluated on our benchmark (\cref{sec:bm_ffnvs}).

As reported in \cref{tab:ff3dgs_bm_ctx300}, all models perform substantially better on DL3DV than on the other datasets, suggesting that 3DGS-based NVS is sensitive to trajectory and pose distributions standardized by DL3DV, rather than scene content. Comparing the two groups, geometry-model-based frameworks consistently outperform specialized feed-forward models, demonstrating that a simple backbone plus DPT head can surpass complex task-specific designs. The advantage stems from large-scale pretraining, which enables better generalization and scalability than approaches relying on epipolar transformers, cost volumes, or cascaded modules.
Within this group, NVS performance correlates with geometry estimation capability, making DA3 the strongest backbone. 
Looking forward, we expect FF-NVS can be effectively addressed with simple architectures leveraging pretrained geometry backbones, and that the strong spatial understanding of DA3 will benefit other 3D vision tasks.

\subsection{Analysis for Depth Anything 3}
Training our DA3-Giant model requires 128$\times$H100 GPUs for approximately 10 days. To reduce carbon footprint and computational cost, all ablation experiments reported in this section are conducted using the ViT-L backbone with a maximum of 10 views, requiring approximately 4 days on 32$\times$H100 GPUs.

\begin{table}[t]
    \caption{
        \textbf{Comparisons with SOTA methods on NVS task.} 
    We report NVS comparsions with exisiting feed-forward 3DGS models and counterparts using other backbones. For each scene, we use 12 input context views and test on target views sampled every 8 views over a set of over 300 views. Image resolution is $270\times480$.    }
    \label{tab:ff3dgs_bm_ctx300}
    \begin{center}
    \resizebox{\linewidth}{!}{

    \begin{tabular}{lccccccccc}
        \toprule
        \multirow{3}{*}[-4pt]{\textbf{Methods}} 
        & \multicolumn{3}{c}{\textbf{In-domain Dataset}} 
        & \multicolumn{6}{c}{\textbf{Out-of-domain Datasets}} \\
        \cmidrule(lr){2-4} \cmidrule(lr){5-10}
        & \multicolumn{3}{c}{\textbf{DL3DV-Benchmarks (140)}} 
        & \multicolumn{3}{c}{\textbf{Tanks and Temples (6)}} 
        & \multicolumn{3}{c}{\textbf{MegaDepth (19)}} \\
        \cmidrule(lr){2-4} \cmidrule(lr){5-7} \cmidrule(lr){8-10}
        & PSNR$\uparrow$ & SSIM$\uparrow$ & LPIPS$\downarrow$ 
        & PSNR$\uparrow$ & SSIM$\uparrow$ & LPIPS$\downarrow$ 
        & PSNR$\uparrow$ & SSIM$\uparrow$ & LPIPS$\downarrow$ \\
        \midrule
        pixelSplat & 16.55 & 0.456 & 0.480 & 13.81 & 0.347 & 0.558 & 13.87 & 0.367 & 0.561 \\
        MVSplat & 18.13 & 0.559 & 0.393 & 14.81 & 0.391 & 0.508 & 14.67 & 0.398 & 0.533 \\ 
        DepthSplat & 19.24 & 0.620 & 0.322&15.80 & 0.474 & 0.418 &15.90 & 0.471 & 0.450 \\  
        \midrule
        Fast3R & 19.30 &  0.604 &  0.320 & 16.24 &  0.478 &  0.409 & 16.43 &  0.493 &  0.421 \\
        MV-DUSt3R & 20.01 &  0.645 &  0.294  & 17.04 &  0.529 &  0.370  & 16.20 &  0.484 &  0.437 \\
        VGGT & 20.96 & 0.697 & 0.253 &17.18 & 0.550 & 0.347 &16.45 & 0.500 & 0.417 \\
        DAv3 (Ours) & \textbf{21.33} &  \textbf{0.711} &  \textbf{0.241} & \textbf{18.10} &  \textbf{0.578} &  \textbf{0.311} & \textbf{17.89} &  \textbf{0.561} &  \textbf{0.351} \\
        \bottomrule
        
    \end{tabular}

    }
    \end{center}
\end{table}

\subsubsection{Sufficiency of the Depth-Ray Representation}

To validate our depth-ray representation, we compare different prediction combinations summarized in ~\cref{tab:ablation_minimal_targets}. 
All models use a ViT-L backbone, identical training settings (view size: 10, batch size: 128, steps: 120k).
We evaluate four heads: 1) \textbf{\depth} for dense depth maps; 2) \textbf{\pcd} for direct 3D point clouds; 3) \textbf{\cam} for 9-DoF camera pose $\mathbf{c}=(\mathbf{t}, \mathbf{q}, \mathbf{f})$; and 4) our proposed \textbf{\ray}, predicting per-pixel ray maps (\cref{sec:formulation}). 
The \ray head uses a Dual-DPT architecture, while \pcd uses a separate DPT head. For models without \pcd, point clouds are obtained by combining \depth with camera parameters from \ray or \cam. 
As shown in Table~\ref{tab:ablation_minimal_targets}, the minimal \depth + \ray configuration consistently outperforms \depth + \pcd + \cam and \depth + \cam across all datasets and metrics, achieving nearly 100\% relative gain in Auc3 over \depth + \cam. 
Adding an auxiliary \cam head (\depth + \ray + \cam) yields no further benefit, confirming the sufficiency of the depth-ray representation.
We adopt \depth + \ray + \cam as our final representation, as the camera head incurs negligible computational overhead, amounting to approximately 0.1\% of the computation cost of the main backbone.

\begin{table}[t]
    \caption{
        \textbf{Ablations of prediction-target combinations.} 
        Note that all experiments in this table do not have camera condition token.
        The \textbf{best} and \underline{second best} are highlighted.
    }
    \small
    \begin{center}
    \resizebox{0.925\linewidth}{!}{
    \begin{tabular}{lcccccccccc}
        \toprule
        \multirow{2}{*}[-4pt]{\textbf{Methods}} 
        & \multicolumn{2}{c}{\textbf{ \picodataset }} 
        & \multicolumn{2}{c}{\textbf{ ETH3D }} 
        & \multicolumn{2}{c}{\textbf{ DTU }} 
        & \multicolumn{2}{c}{\textbf{ 7Scenes }} 
        & \multicolumn{2}{c}{\textbf{ ScanNet++ }} 
        \\
        \cmidrule(lr){2-3} \cmidrule(lr){4-5} \cmidrule(lr){6-7} \cmidrule(lr){8-9} \cmidrule(lr){10-11} 
        & Auc3$\uparrow$ & F1$\uparrow$ 
        & Auc3$\uparrow$ & F1$\uparrow$ 
        & Auc3$\uparrow$ & CD$\downarrow$ 
        & Auc3$\uparrow$ & F1$\uparrow$ 
        & Auc3$\uparrow$ & F1$\uparrow$  \\
        \midrule
        depth + pcd + cam   & 9.1 & 12.8 & 19.0 & \underline{60.4} & 42.3 & 4.918 & 20.8 & 43.4 & 22.0 & 43.0\\
        depth + cam         & 10.8 & 16.5 & 9.9 & 48.0 & 23.3 & 5.316 & 13.0 & 38.5 & 13.3 & 41.0\\
        depth + ray         & \textbf{48.7} & \textbf{60.3} & \textbf{25.5} & \textbf{65.4} & \underline{46.5} & \underline{3.919} & \underline{24.0} & \textbf{46.5} & \textbf{35.5} & \underline{53.4}\\ 
        \depth + \ray + \cam   & \underline{37.2} & \underline{45.4} & \underline{22.3} & 59.4 & \textbf{56.3} & \textbf{3.066} & \textbf{25.7} & \underline{45.6} & \underline{34.1} & \textbf{56.5}\\ 
        \bottomrule
    \end{tabular}
    }
    \end{center}
    \label{tab:ablation_minimal_targets}
\end{table}
\begin{table}[tb]
    \caption{
        \textbf{Ablation study.} We evaluate three architectural designs with comparable model sizes (a-c), the effects of the dual-DPT head (d), teacher label supervision (e), and the pose conditioning module (f-g). The \textbf{best} and \underline{second best} are highlighted. Methods marked with "*" are evaluated with ground-truth pose fusion.
    }
    \label{tab:abl_dav3}
    \begin{center}
    \resizebox{0.925\linewidth}{!}{
    \begin{tabular}{lcccccccccc}
        \toprule
        \multirow{2}{*}[-4pt]{\textbf{Methods}} 
        & \multicolumn{2}{c}{\textbf{ \picodataset }} 
        & \multicolumn{2}{c}{\textbf{ ETH3D }} 
        & \multicolumn{2}{c}{\textbf{ DTU }} 
        & \multicolumn{2}{c}{\textbf{ 7Scenes }} 
        & \multicolumn{2}{c}{\textbf{ ScanNet++ }} 
        \\
        \cmidrule(lr){2-3} \cmidrule(lr){4-5} \cmidrule(lr){6-7} \cmidrule(lr){8-9} \cmidrule(lr){10-11} 
        & Auc3$\uparrow$ & F1$\uparrow$ 
        & Auc3$\uparrow$ & F1$\uparrow$ 
        & Auc3$\uparrow$ & CD$\downarrow$ 
        & Auc3$\uparrow$ & F1$\uparrow$ 
        & Auc3$\uparrow$ & F1$\uparrow$  \\
        \midrule
        a. Proposed Arch.      
        & \textbf{39.2} & \textbf{47.0} 
        & \textbf{21.0} & \textbf{55.4} 
        & \textbf{45.8} & \textbf{3.82} 
        & \textbf{26.2} & \underline{47.6} 
        & \textbf{30.3} & \textbf{51.1} \\
        b. VGGT Style      
        & 3.72 & 14.5 
        & 2.31 & 27.4 
        & 1.38 & 6.93 
        & 0.97 & 21.4 
        & 2.03 & 12.2 \\
        c. Full Alt.       
        & \underline{24.7} & \underline{29.3} 
        & \underline{13.1} & \underline{51.9} 
        & \underline{44.6} & \underline{4.23} 
        & \underline{21.1} & \textbf{48.6} 
        & \underline{27.7} & \underline{47.5} \\
        \midrule
        d. w/o Dual DPT    
        & 5.59 & 11.5 
        & 13.6 & 33.4 
        & 21.7 & 5.14 
        & 14.2 & 49.4 
        & 26.5 & 46.6 \\
        e. w/o Teacher     
        & 11.2 & 16.0 
        & 16.2 & 57.6 
        & 52.5 & 3.29 
        & 23.3 & 40.3 
        & 26.2 & 47.7 \\
        \midrule
        f. w/o Pose Cond.*
        &  & 65.8
        &  & 63.2
        &  & 3.65
        &  & 58.4
        &  & 62.8 \\
        g. w/ Pose Cond.*
        &  & 73.8
        &  & 70.9
        &  & 2.14
        &  & 46.0
        &  & 65.7 \\
        \bottomrule
    \end{tabular}
    }
    \end{center}
\end{table}

\subsubsection{Sufficiency of a Single Plain Transformer}
We compare a standard ViT-L backbone with a VGGT-style architecture that stacks two distinct transformers, tripling the block count. For fair capacity comparison, the VGGT-style model uses smaller ViT-B backbones, yielding a similar parameter size to our ViT-L. Our backbone supports two attention strategies: \textbf{Full Alt.}, which alternates cross-view/within-view attention in all layers ($L=L_\text{g}$), and our default partial alternation.
As shown in Table~\ref{tab:abl_dav3}, the VGGT-style model drops to 79.8\% of our baseline performance, confirming the superiority of a single-transformer design at similar scale. We attribute this gap to full pretraining of our backbone versus two-thirds untrained blocks in VGGT. Moreover, the \textbf{Full Alt.} variant degrades across nearly all metrics—except F1 on 7Scenes—indicating that partial alternation is the more effective and robust strategy.

\subsubsection{Ablation and Analysis}

\paragraph{Dual-DPT Head.}
We assess the effectiveness of the dual-DPT head via an ablation in which two separate DPT heads predict depth and ray maps independently. Results are reported in \cref{tab:abl_dav3}, item (d). Compared with the model equipped with the dual-DPT head, the variant without it shows consistent drops across metrics, confirming the effectiveness of our dual-DPT design.

\paragraph{Teacher model supervision.}
We ablate the use of teacher model labels as supervision, with quantitative results reported in \cref{tab:abl_dav3}, item (e). Training without teacher labels yields a slight improvement on DTU but leads to performance drops on 7Scenes and ScanNet++. Notably, the degradation is pronounced on HiRoom. We attribute this to HiRoom's synthetic nature and its ground truth containing abundant fine structures; supervision from the teacher helps the student capture such details more accurately. Qualitative comparisons in \cref{fig:vis_teacher_sup} corroborate this trend: models trained with teacher-label supervision produce depth maps with substantially richer detail and finer structures.
\paragraph{Pose conditioning.}
To assess the pose-conditioning module, we ablate it on the ViT-L backbone and report results in \cref{tab:abl_dav3}, items (f) and (g). Unlike other entries in the table, these two are evaluated with ground-truth pose fusion (marked with ``*''), whereas the rest use predicted pose fusion. Across metrics, configurations with pose conditioning consistently outperform those without, confirming the effectiveness of the pose-conditioning module.

\paragraph{Running time.} 
We present analysis on Parameters, max number of images and running speed in ~\cref{tab:analysis}
\begin{table}[t]
    \caption{
        \textbf{Comparison of Models with Parameters and Running Speed.} 
        The maximum number of images was tested on an 80 GB A100 GPU.
        If we store some intermediate tokens in CPU memory, we could process many more images.
        The running speed was measured on an A100 GPU with a scene of 32 images, and we report the average speed per image.
        The image resolution is $504\times336$.
    }
    \small
    \begin{center}
    \resizebox{0.95\linewidth}{!}{
    \begin{tabular}{l c ccc c}
        \toprule
        \multirow{2}{*}[-4pt]{\textbf{Model}} 
        & \multirow{2}{*}[-4pt]{\textbf{Max \# of Images}} 
        & \multicolumn{3}{c}{\textbf{Parameters}} 
        & \multirow{2}{*}[-4pt]{\textbf{Running Speed}} \\
        \cmidrule(lr){3-5} 
        &  & Backbone & DualDPT & CameraHead &  \\
        \midrule
        VGGT(Reference) & 400-500 & 0.91B   & 0.064B & 0.22B  & 34.1 FPS \\
        \midrule
        DA3-Giant & 900-1000 & 1.130B   & 0.050B & 0.018B  & 37.6 FPS \\
        DA3-Large & 1500-1600  & 0.300B    & 0.047B  & 0.008B  & 78.37 FPS  \\
        DA3-Base & 2100-2200  &  0.086B   & 0.015B  & 0.004B  & 126.5 FPS  \\
        DA3-Small & 4000-4100 & 0.022B      & 0.003B  & 0.001B  & 160.5 FPS  \\
        \bottomrule
    \end{tabular}
    }
    \end{center}
    \label{tab:analysis}
\end{table}

\begin{figure}[h]
    \begin{center}
        \includegraphics[width=1\textwidth]{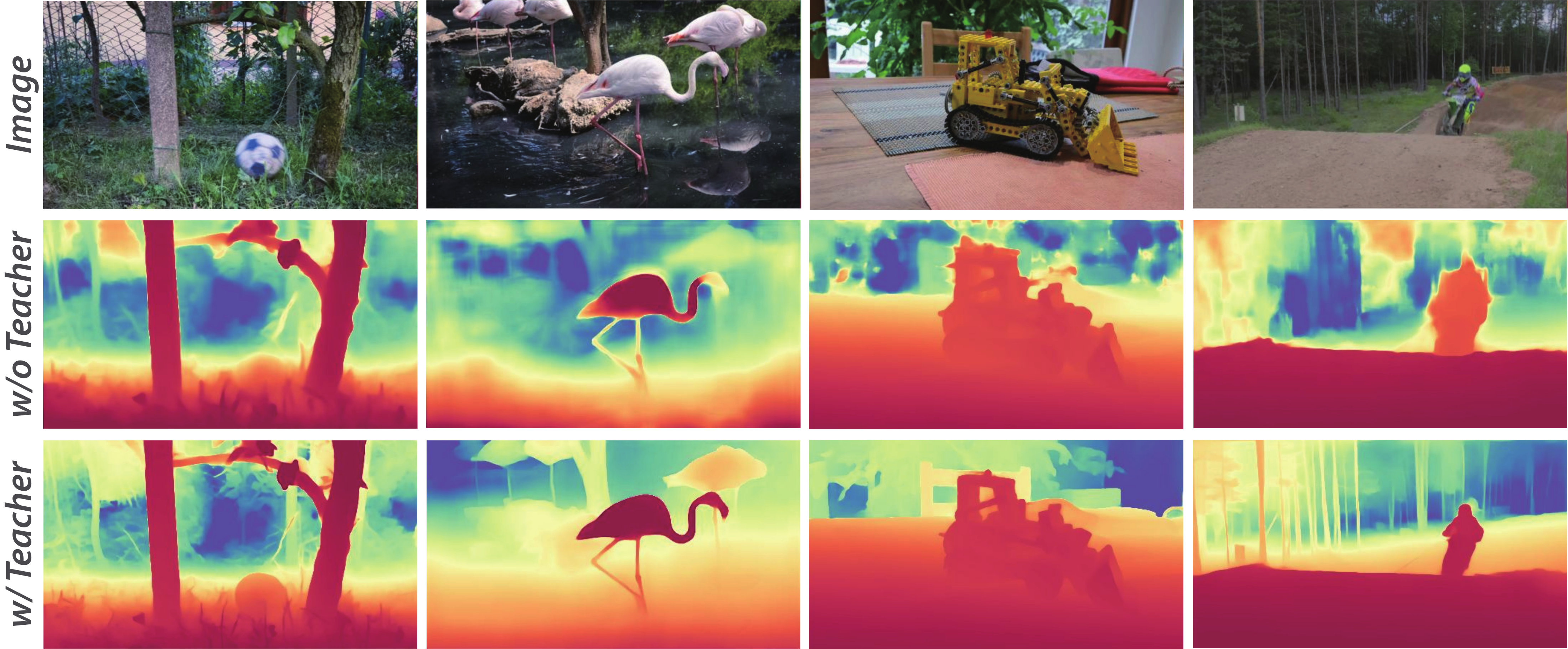}
        \caption{\textbf{Comparison of teacher-label supervision.} Supervision with teacher-generated labels yields depth maps with substantially richer detail and finer structures.
        }
        \label{fig:vis_teacher_sup}
    \end{center}
\end{figure}

\begin{table}[t]
\centering
\caption{\textbf{Ablations for teacher model.} 
Training with V3 datasets and multi-resolution strategy yields the best performance. Depth-based geometry achieves the best AbsRel and SqRel. The full teacher-loss outperforms other variants.
(AbsRel: $\downarrow$, SqRel: $\downarrow$, $\delta_1$: $\uparrow$). The results are averaged over KITTI, NYU, ETH3D, SUN-RGBD and DIODE.
}
\vspace{-3mm}
\resizebox{\linewidth}{!}{
\begin{tabular}{lccc|lccc|lccc}
\toprule
\multicolumn{4}{c|}{\textbf{Data}} & \multicolumn{4}{c|}{\textbf{Geometry}} & \multicolumn{4}{c}{\textbf{Loss}} \\
\midrule
Data & $\delta_1$& AbsRel & SqRel & Geometry & $\delta_1$& AbsRel & SqRel & Loss & $\delta_1$& AbsRel & SqRel \\
\midrule 
V2 & 0.919  & 0.087  & 0.596 & Disparity & \textbf{0.919}  & \underline{0.095}  & 1.033 & MAE-Loss & 0.918  & 0.089  & 0.637 \\
V3 & \underline{0.929}  & \underline{0.079}  & \underline{0.508} & Pointmap & 0.912  & 0.096  & \underline{0.693} & w/o Dist. Nor. & 0.918  & \textbf{0.087}  & \underline{0.600} \\
\textbf{V3 + mr.} & \textbf{0.938}  & \textbf{0.072}  & \textbf{0.452} & \textbf{Depth} & \underline{0.918}  & \textbf{0.089}  & \textbf{0.637} & \textbf{Full loss} & \textbf{0.919}  & \textbf{0.087}  & \textbf{0.596} \\
\bottomrule
\end{tabular}}
\label{tab:ablation_teacher}
\end{table}

\begin{figure}[h]
    \vspace{-1mm}
    \begin{center}
        \includegraphics[width=1\textwidth]{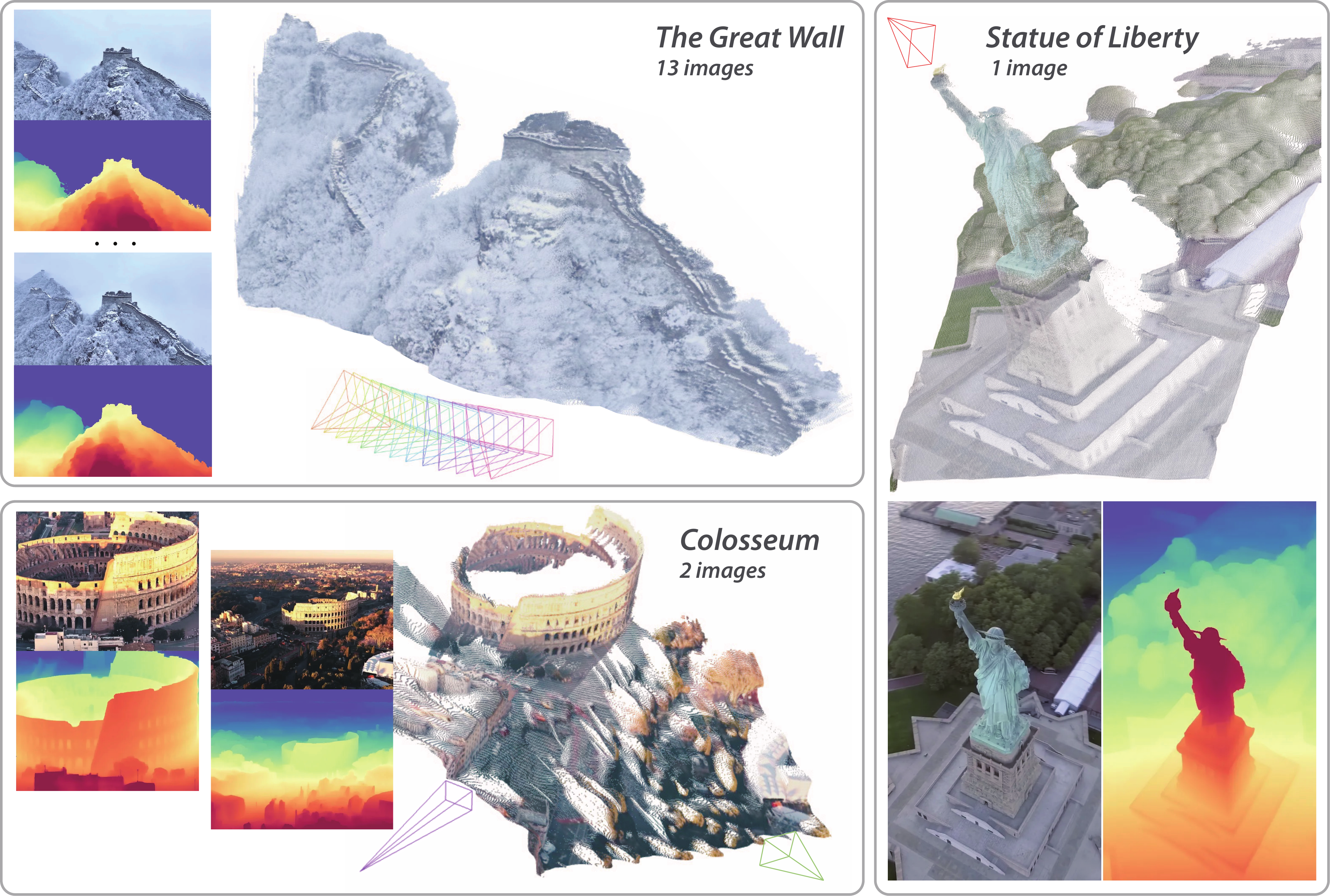}
    \vspace{-3mm}
        \caption{\textbf{Visualizations of camera pose and depth estimation on in-the-wild scenes.} 
        }
    \vspace{-5mm}
        \label{fig:vis_depth_wild}
    \end{center}
\end{figure}

\paragraph{More visualizations.}
We provide additional visualizations of camera pose and depth estimation on in-the-wild scenes in \cref{fig:vis_depth_wild}, demonstrating the robustness and quality of our model across diverse real-world scenarios.

\subsection{Analysis for Depth-Anything-3-Monocular}

\subsubsection{Teacher Model}
The teacher model’s metrics are reported in \cref{tab:monodepth}. Our new teacher consistently outperforms DA2 across all datasets, with the sole exception of NYU, where performance is on par with DA2.
For the teacher ablation, we employ a ViT-L backbone and a batch size of 64. Evaluation follows the DA2 benchmark protocol, and we additionally report Squared Relative Error (SqRel), defined as the mean squared error between predictions and ground truth normalized by the ground truth.
As shown in \cref{tab:ablation_teacher}, across geometries, depth emerges as the most effective target compared with disparity and point maps. For training objectives, the full teacher loss proposed in this work outperforms both the DA2 loss and a variant without proposed normal-loss term. Finally, data scaling contribute notably to performance: upgrading datasets from V2 to V3 and adopting a multi-resolution training strategy yield consistent improvements in the teacher's final metrics.

\subsubsection{Student Model}
As shown in \cref{tab:student_mono}, our monocular student model with a ViT-L backbone outperforms the DA2 student across all evaluation datasets. Notably, on the ETH3D~\cite{schops2017eth3d} benchmark the new monocular student achieves an improvement of over $10\%$ compared with DA2.
The improved performance is attributed to the enhanced teacher model with better geometry supervision and the scaled training data (V3). On challenging datasets like SINTEL, our student also demonstrates substantial gains (+5.1\%), validating the effectiveness of our teacher-student distillation framework.

\begin{table}[b!]
    \centering
    \caption{\textbf{Monocular student depth comparisons.} $\delta_1 \uparrow$}
    \label{tab:student_mono}
    \resizebox{0.6\linewidth}{!}{
    \begin{tabular}{lccccc}
        \toprule
        Method & KITTI & NYU & SINTEL & ETH3D & DIODE\\
        \midrule
        DA2 & 94.6 & 97.9 & 77.2 & 86.5 & 95.2\\
        mono-student & 97.1 & 98.0 & 82.3 & 98.8 & 96.5\\
        \bottomrule
    \end{tabular}
    }
\end{table}

\subsection{Analysis for Depth-Anything-3-Metric}

We compare with state-of-the-art metric depth estimation methods, including DepthPro~\cite{bochkovskiydepth}, Metric3D v2~\cite{hu2024metric3dv2}, UniDepthv1~\cite{piccinelli2024unidepth} and UniDepthv2~\cite{piccinelli2025unidepthv2}, on 5 benchmarks: NYUv2~\cite{silberman2012nyu}, KITTI~\cite{geiger2013kitti}, ETH3D~\cite{schops2017eth3d}, SUN-RGBD~\cite{song2015sunrgbd} and DIODE (indoor)~\cite{vasiljevic2019diode}. 

As shown in \cref{tab:metric}, DA3-metric achieves \textbf{state-of-the-art performance on ETH3D} ($\delta_1$ = 0.917, AbsRel = 0.104), substantially outperforming the second-best method UniDepthv2 ($\delta_1$ = 0.863) by a large margin. DA3-metric also achieves \textbf{best performance on SUN-RGBD} for AbsRel (0.105) and \textbf{second-best on DIODE} ($\delta_1$ = 0.838, AbsRel = 0.128). While UniDepthv1 and UniDepthv2 achieve the best results on NYUv2 and KITTI, DA3-metric demonstrates strong generalization and competitive performance across all benchmarks, particularly excelling on diverse outdoor scenes like ETH3D.

We ablate the Teacher supervision in \cref{tab:metric}. The results show interesting trade-offs: removing Teacher supervision slightly improves metrics on NYUv2 and KITTI, while maintaining comparable performance on other datasets. As shown in Fig.~\ref{fig:metric-abl}, Teacher supervision significantly improves sharpness and fine detail quality, demonstrating that Teacher provides complementary knowledge beyond standard metrics.

\begin{table}[tb]
    \caption{\textbf{Comparison with state-of-the-arts on metric depth estimation.} 
    The \textbf{best} and \underline{second best} are highlighted. Bottom rows show ablation results with and without teacher supervision. 
    Note that the ablation setting is slightly different from the final model on training resolution, which leads to minor differences in performance.}
    \label{tab:metric}
    \centering
    \resizebox{\linewidth}{!}{
    \begin{tabular}{lcccccccccc}
        \toprule
        \multirow{2}{*}[-4pt]{\textbf{Methods}} 
        & \multicolumn{2}{c}{\textbf{ NYUv2 }} 
        & \multicolumn{2}{c}{\textbf{ KITTI }} 
        & \multicolumn{2}{c}{\textbf{ ETH3D }} 
        & \multicolumn{2}{c}{\textbf{ SUN-RGBD }} 
        & \multicolumn{2}{c}{\textbf{ DIODE }} 
        \\
        \cmidrule(lr){2-3} \cmidrule(lr){4-5} \cmidrule(lr){6-7} \cmidrule(lr){8-9} \cmidrule(lr){10-11} 
        & $\delta_1$$\uparrow$ & AbsRel$\downarrow$ 
        & $\delta_1$$\uparrow$ & AbsRel$\downarrow$ 
        & $\delta_1$$\uparrow$ & AbsRel$\downarrow$ 
        & $\delta_1$$\uparrow$ & AbsRel$\downarrow$ 
        & $\delta_1$$\uparrow$ & AbsRel$\downarrow$  \\
        \midrule
        DepthPro~\cite{bochkovskiydepth} & 0.932 & 0.093 & 0.843 & 0.121 & 0.386 & 0.349 & 0.950 & 0.126 & 0.734 & 0.173 \\
        Metric3D v2~\cite{hu2024metric3d} & \underline{0.971} & 0.067 & \underline{0.976} & \textbf{0.051} & 0.830 & \underline{0.138} & 0.954 & 0.132 & 0.018 & 0.154 \\
        UniDepthv1~\cite{piccinelli2024unidepth} & \textbf{0.980} & \textbf{0.061} & \textbf{0.978} & \textbf{0.051}
        & 0.234 & 0.464 & \underline{0.971} & 0.113 & 0.570 & 0.266 \\
        UniDepthv2~\cite{piccinelli2025unidepthv2} & 0.968 & \underline{0.064} & 0.968 & 0.076 & \underline{0.863} & 0.152 & \textbf{0.977} & \underline{0.111} & \textbf{0.856} & \textbf{0.123} \\
        \midrule
        DA3-metric & 0.963 & 0.070 & 0.953 & 0.086 & \textbf{0.917} & \textbf{0.104} & \underline{0.973} & \textbf{0.105} & \underline{0.838} & \underline{0.128} \\  
        \midrule
        w/ teacher & 0.966 & 0.073 & 0.947 & 0.086 & 0.906 & 0.105 & 0.973 & 0.104 & 0.824 & 0.132 \\
        w/o teacher & 0.969 & 0.066 & 0.965 & 0.067 & 0.907 & 0.105 & 0.975 & 0.099 & 0.816 & 0.134 \\
        \bottomrule
    \end{tabular}
    }
\end{table}

\begin{figure}[h]
    \centering
    \includegraphics[width=\linewidth]{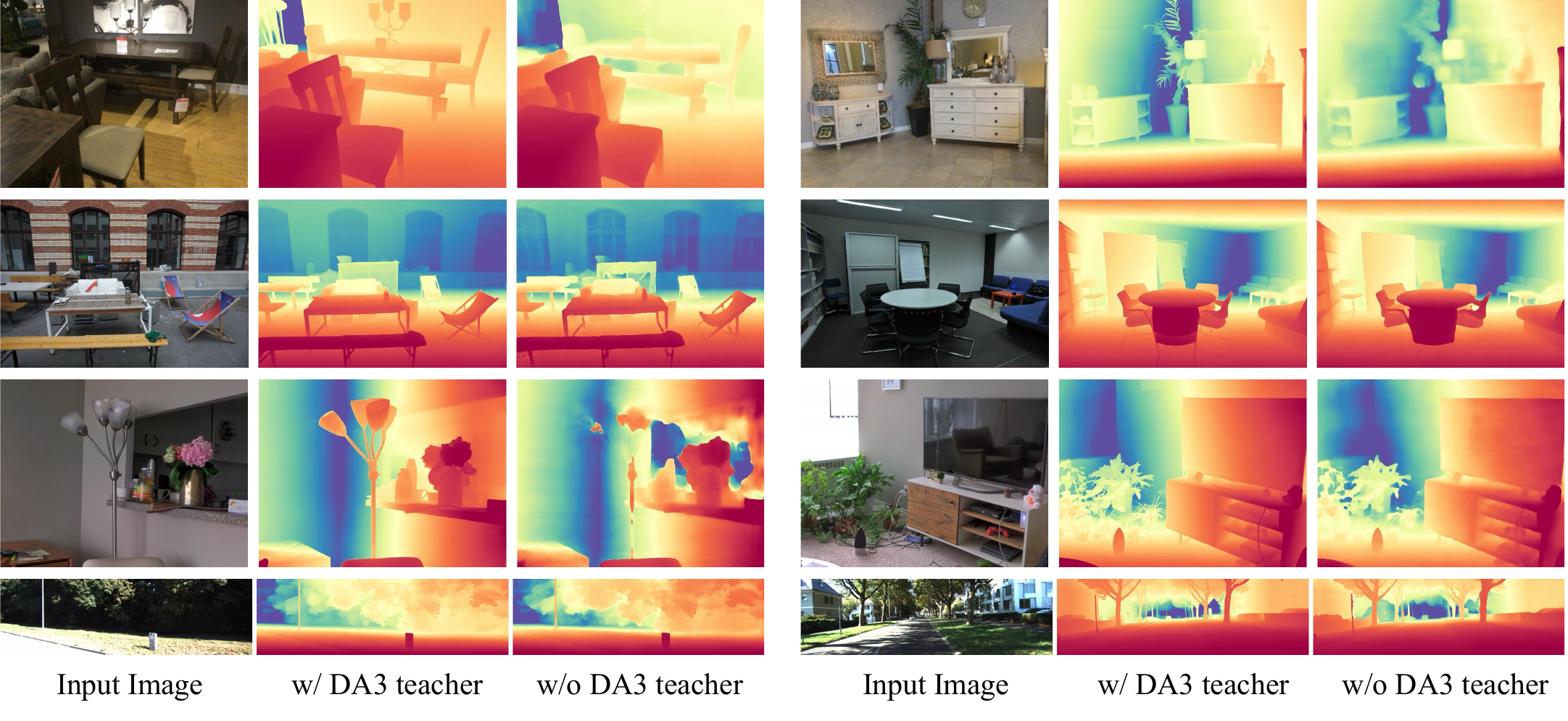}
    \caption{\textbf{Effectiveness of Teacher model for supervising metric depth estimation.} Incorporating Teacher model for supervision significantly improves the metric depth sharpness.}
    \label{fig:metric-abl}
\end{figure}

\begin{figure}[h]
    \begin{center}
        \vspace{-1mm}
        \includegraphics[width=0.8\textwidth]{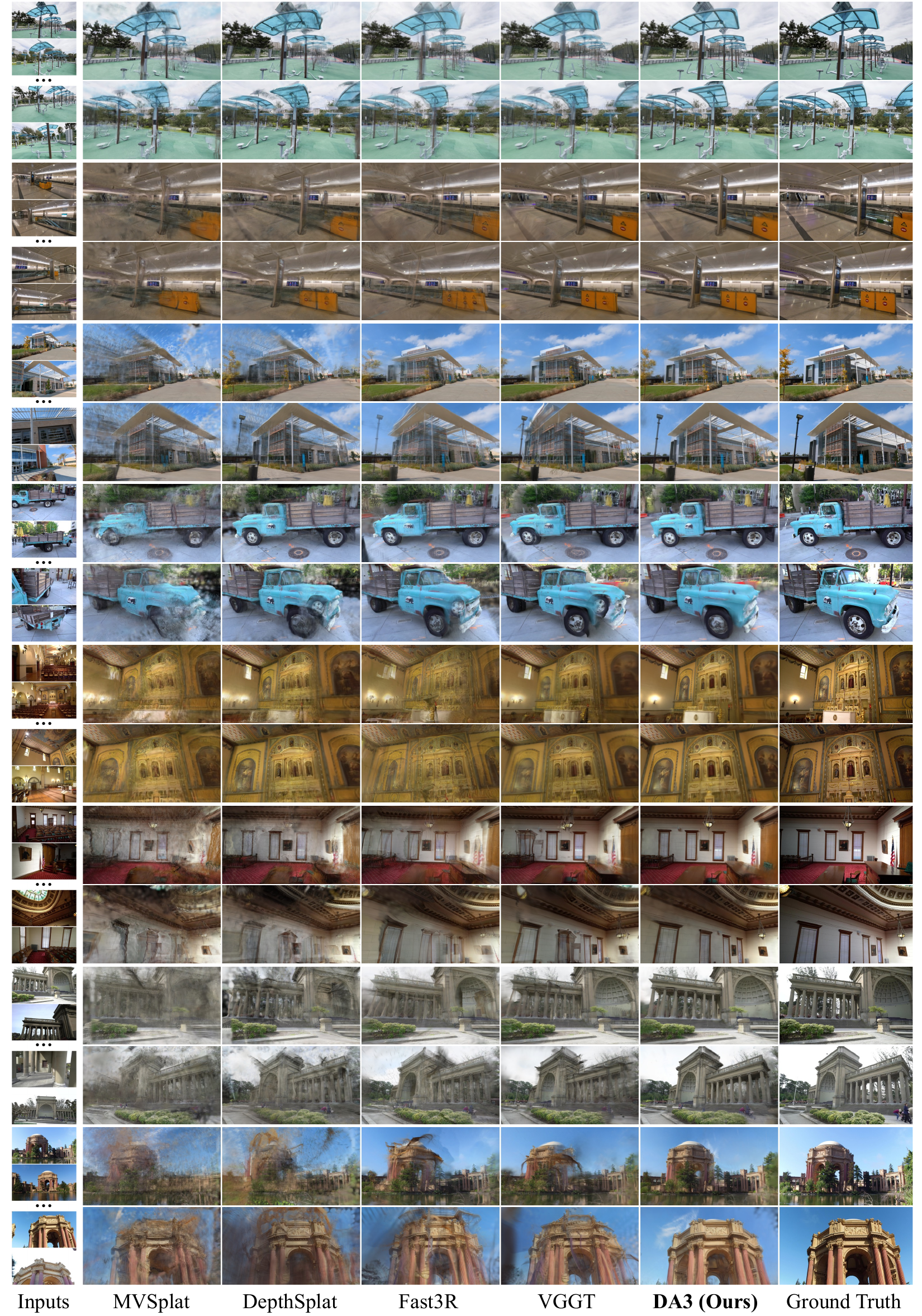}
        \caption{
        \textbf{Qualitative comparisons with state-of-the-art methods for visual rendering.} The first column shows the selected input views, while the remaining columns display novel views rendered by comparison models and ground truth. For each scene, two rendered novel viewpoints are presented in consecutive rows. The first three scenes are from DL3DV, the following two are from Tanks and Temples, and the last three are from MegaDepth. Compared to other methods, our model consistently achieves superior rendering quality across diverse and challenging scenes.
        }
        \label{fig:vis_nvs_render}
    \end{center}
\end{figure}

\subsection{Analysis for Feed-forward 3DGS}

We retrain all compared feed-forward 3DGS models, ensuring that the training configuration matches the testing setup by using 12 input context views selected through farthest point sampling. We apply engineering optimizations such as flash attention and fully shared data parallelism to enable all models to process 12 input views efficiently. Depth training loss are incorporated for all baselines to ensure stable training and convergence. All models are trained on 8 A100 GPUs for 200K steps with a batch size of 1, except for pixelSplat, which is trained for 100K steps due to rather slow epipolar attention. All results are reported at $H \times W = 270 \times 480$.

\paragraph{Visual quality analysis.}
We present visual comparisons with other models in \cref{fig:vis_nvs_render} under novel view synthesis settings. As illustrated, simply augmenting our DA3 model with a 3D Gaussian DPT head yields significantly improved rendering quality over existing state-of-the-art approaches. Our model demonstrates particular strength in challenging regions, such as thin structures (\eg, columns in the first and third scenes) and large-scale outdoor environments with wide-baseline input views (last two scenes), as shown in \cref{fig:vis_nvs_render}. These results underscore the importance of a robust geometry backbone for high-quality visual rendering, consistent with our quantitative findings in \cref{tab:ff3dgs_bm_ctx300}. We anticipate that the strong geometric understanding of DA3 will also benefit other 3D vision tasks.

\section{Conclusion and Discussion}

Depth Anything 3 shows that a plain transformer, trained on depth-and-ray targets with teacher--student supervision, can unify any-view geometry without ornate architectures. Scale-aware depth, per-pixel rays, and adaptive cross-view attention let the model inherit strong pretrained features while remaining lightweight and easy to extend.
On the proposed visual geometry benchmark the approach sets new pose and reconstruction records, with both giant and compact variants surpassing prior models, while the same backbone powers efficient feed-forward novel view synthesis model.

We view Depth Anything 3 as a step toward versatile 3D foundation models. Future work can extend its reasoning to dynamic scenes, integrate language and interaction cues, and explore larger-scale pretraining to close the loop between geometry understanding and actionable world models. We hope the model and dataset releases, benchmark, and simple modeling principles offered here catalyze broader research on general-purpose 3D perception.

\section*{Acknowledgement}
We thank Xiaowei Zhou, Sida Peng and Hengkai Guo for their valuable discussions during the development of this project.
We are also grateful to Yang Zhao for his engineering support.
The input images in the teaser demo were extracted from a publicly available YouTube video \citep{mtsdrones2024}, credited to the original creator.

\clearpage

\bibliographystyle{plainnat}
\bibliography{main}

@String(PAMI = {IEEE Trans. Pattern Anal. Mach. Intell.})

@String(IJCV = {Int. J. Comput. Vis.})

@String(CVPR= {IEEE Conf. Comput. Vis. Pattern Recog.})

@String(ICCV= {Int. Conf. Comput. Vis.})

@String(ECCV= {Eur. Conf. Comput. Vis.})

@String(NIPS= {Adv. Neural Inform. Process. Syst.})

@String(TOG= {ACM Trans. Graph.})

@String(ICLR = {Int. Conf. Learn. Represent.})

@article{arterberry2000perception,
  title={Perception of three-dimensional shape specified by optic flow by 8-week-old infants},
  author={Arterberry, Martha E and Yonas, Albert},
  journal={Perception \& Psychophysics},
  volume={62},
  number={3},
  pages={550--556},
  year={2000},
  publisher={Springer}
}

@article{yang2024depth,
  title={Depth anything v2},
  author={Yang, Lihe and Kang, Bingyi and Huang, Zilong and Zhao, Zhen and Xu, Xiaogang and Feng, Jiashi and Zhao, Hengshuang},
  journal=NIPS,
  volume={37},
  pages={21875--21911},
  year={2024}
}

@article{oquab2023dinov2,
  title={Dinov2: Learning robust visual features without supervision},
  author={Oquab, Maxime and Darcet, Timoth{\'e}e and Moutakanni, Th{\'e}o and Vo, Huy and Szafraniec, Marc and Khalidov, Vasil and Fernandez, Pierre and Haziza, Daniel and Massa, Francisco and El-Nouby, Alaaeldin and others},
  journal={arXiv preprint arXiv:2304.07193},
  year={2023}
}

@inproceedings{ranftl2021vision,
  title={Vision transformers for dense prediction},
  author={Ranftl, Ren{\'e} and Bochkovskiy, Alexey and Koltun, Vladlen},
  booktitle=ICCV,
  pages={12179--12188},
  year={2021}
}

@inproceedings{wang2024dust3r,
  title={Dust3r: Geometric 3d vision made easy},
  author={Wang, Shuzhe and Leroy, Vincent and Cabon, Yohann and Chidlovskii, Boris and Revaud, Jerome},
  booktitle=CVPR,
  pages={20697--20709},
  year={2024}
}

@inproceedings{yang2025fast3r,
  title={Fast3r: Towards 3d reconstruction of 1000+ images in one forward pass},
  author={Yang, Jianing and Sax, Alexander and Liang, Kevin J and Henaff, Mikael and Tang, Hao and Cao, Ang and Chai, Joyce and Meier, Franziska and Feiszli, Matt},
  booktitle=CVPR,
  pages={21924--21935},
  year={2025}
}

@inproceedings{wang2025vggt,
  title={Vggt: Visual geometry grounded transformer},
  author={Wang, Jianyuan and Chen, Minghao and Karaev, Nikita and Vedaldi, Andrea and Rupprecht, Christian and Novotny, David},
  booktitle=CVPR,
  pages={5294--5306},
  year={2025}
}

@inproceedings{silberman2012nyu,
  title={Indoor segmentation and support inference from rgbd images},
  author={Silberman, Nathan and Hoiem, Derek and Kohli, Pushmeet and Fergus, Rob},
  booktitle=ECCV,
  pages={746--760},
  year={2012},
  organization={Springer}
}

@article{geiger2013kitti,
  title={Vision meets robotics: The kitti dataset},
  author={Geiger, Andreas and Lenz, Philip and Stiller, Christoph and Urtasun, Raquel},
  journal={The international journal of robotics research},
  volume={32},
  number={11},
  pages={1231--1237},
  year={2013},
  publisher={Sage Publications Sage UK: London, England}
}

@inproceedings{schops2017eth3d,
  title={A multi-view stereo benchmark with high-resolution images and multi-camera videos},
  author={Schops, Thomas and Schonberger, Johannes L and Galliani, Silvano and Sattler, Torsten and Schindler, Konrad and Pollefeys, Marc and Geiger, Andreas},
  booktitle=cvpr,
  pages={3260--3269},
  year={2017}
}

@inproceedings{song2015sunrgbd,
  title={Sun rgb-d: A rgb-d scene understanding benchmark suite},
  author={Song, Shuran and Lichtenberg, Samuel P and Xiao, Jianxiong},
  booktitle=cvpr,
  pages={567--576},
  year={2015}
}

@article{vasiljevic2019diode,
  title={Diode: A dense indoor and outdoor depth dataset},
  author={Vasiljevic, Igor and Kolkin, Nick and Zhang, Shanyi and Luo, Ruotian and Wang, Haochen and Dai, Falcon Z and Daniele, Andrea F and Mostajabi, Mohammadreza and Basart, Steven and Walter, Matthew R and others},
  journal={arXiv preprint arXiv:1908.00463},
  year={2019}
}

@article{aanaes2016dtu,
  title={Large-scale data for multiple-view stereopsis},
  author={Aan{\ae}s, Henrik and Jensen, Rasmus Ramsb{\o}l and Vogiatzis, George and Tola, Engin and Dahl, Anders Bjorholm},
  journal=IJCV,
  volume={120},
  number={2},
  pages={153--168},
  year={2016},
  publisher={Springer}
}

@inproceedings{shotton2013sevenscene,
  title={Scene coordinate regression forests for camera relocalization in RGB-D images},
  author={Shotton, Jamie and Glocker, Ben and Zach, Christopher and Izadi, Shahram and Criminisi, Antonio and Fitzgibbon, Andrew},
  booktitle=cvpr,
  pages={2930--2937},
  year={2013}
}

@inproceedings{yeshwanth2023scannet++,
  title={Scannet++: A high-fidelity dataset of 3d indoor scenes},
  author={Yeshwanth, Chandan and Liu, Yueh-Cheng and Nie{\ss}ner, Matthias and Dai, Angela},
  booktitle=iccv,
  pages={12--22},
  year={2023}
}

@inproceedings{yao2018mvsnet,
  title={Mvsnet: Depth inference for unstructured multi-view stereo},
  author={Yao, Yao and Luo, Zixin and Li, Shiwei and Fang, Tian and Quan, Long},
  booktitle=eccv,
  pages={767--783},
  year={2018}
}

@article{knapitsch2017tnt,
  title={Tanks and temples: Benchmarking large-scale scene reconstruction},
  author={Knapitsch, Arno and Park, Jaesik and Zhou, Qian-Yi and Koltun, Vladlen},
  journal=tog,
  volume={36},
  number={4},
  pages={1--13},
  year={2017},
  publisher={ACM New York, NY, USA}
}

@inproceedings{mildenhall2020nerf,
 author = {Mildenhall, Ben and Srinivasan, Pratul P and Tancik, Matthew and Barron, Jonathan T and Ramamoorthi, Ravi and Ng, Ren},
 booktitle = ECCV,
 title = {Nerf: Representing scenes as neural radiance fields for view synthesis},
 year = {2020}
}

@inproceedings{eigen2014depth,
  title={Depth map prediction from a single image using a multi-scale deep network},
  author={Eigen, David and Puhrsch, Christian and Fergus, Rob},
  booktitle=NIPS,
  year={2014}
}

@article{snavely2006photo,
  title={Photo tourism: exploring photo collections in 3D},
  author={Snavely, Noah and Seitz, Steven M and Szeliski, Richard},
  journal=TOG,
  pages={835--846},
  year={2006}
}

@inproceedings{seitz2006comparison,
  title={A comparison and evaluation of multi-view stereo reconstruction algorithms},
  author={Seitz, Steven M and Curless, Brian and Diebel, James and Scharstein, Daniel and Szeliski, Richard},
  booktitle=CVPR,
  volume={1},
  pages={519--528},
  year={2006},
  organization={IEEE}
}

@article{mur2015orb,
  title={ORB-SLAM: A versatile and accurate monocular SLAM system},
  author={Mur-Artal, Raul and Montiel, Jose Maria Martinez and Tardos, Juan D},
  journal={IEEE transactions on robotics},
  volume={31},
  number={5},
  pages={1147--1163},
  year={2015},
  publisher={IEEE}
}

@article{baruch2021arkitscenes,
  title={Arkitscenes: A diverse real-world dataset for 3d indoor scene understanding using mobile rgb-d data},
  author={Baruch, Gilad and Chen, Zhuoyuan and Dehghan, Afshin and Dimry, Tal and Feigin, Yuri and Fu, Peter and Gebauer, Thomas and Joffe, Brandon and Kurz, Daniel and Schwartz, Arik and others},
  journal=NIPS,
  year={2021}
}

@inproceedings{reizenstein2021common,
  title={Common objects in 3d: Large-scale learning and evaluation of real-life 3d category reconstruction},
  author={Reizenstein, Jeremy and Shapovalov, Roman and Henzler, Philipp and Sbordone, Luca and Labatut, Patrick and Novotny, David},
  booktitle=ICCV,
  pages={10901--10911},
  year={2021}
}

@inproceedings{yang2024depthv2,
  title={Depth Anything V2},
  author={Yang, Lihe and Kang, Bingyi and Huang, Zilong and Zhao, Zhen and Xu, Xiaogang and Feng, Jiashi and Zhao, Hengshuang},
  booktitle=NIPS,
  year={2024}
}

@inproceedings{xu2025depthsplat,
  title={Depthsplat: Connecting gaussian splatting and depth},
  author={Xu, Haofei and Peng, Songyou and Wang, Fangjinhua and Blum, Hermann and Barath, Daniel and Geiger, Andreas and Pollefeys, Marc},
  booktitle=CVPR,
  pages={16453--16463},
  year={2025}
}

@inproceedings{schoenberger2016sfm,
    author={Sch\"{o}nberger, Johannes Lutz and Frahm, Jan-Michael},
    title={Structure-from-Motion Revisited},
    booktitle=CVPR,
    year={2016},
}

@inproceedings{schoenberger2016mvs,
    author={Sch\"{o}nberger, Johannes Lutz and Zheng, Enliang and Pollefeys, Marc and Frahm, Jan-Michael},
    title={Pixelwise View Selection for Unstructured Multi-View Stereo},
    booktitle=ECCV,
    year={2016},
}

@inproceedings{detone2018superpoint,
  title={Superpoint: Self-supervised interest point detection and description},
  author={DeTone, Daniel and Malisiewicz, Tomasz and Rabinovich, Andrew},
  booktitle={CVPR workshops},
  pages={224--236},
  year={2018}
}

@inproceedings{dusmanu2019d2,
  title={D2-net: A trainable cnn for joint description and detection of local features},
  author={Dusmanu, Mihai and Rocco, Ignacio and Pajdla, Tomas and Pollefeys, Marc and Sivic, Josef and Torii, Akihiko and Sattler, Torsten},
  booktitle=CVPR,
  pages={8092--8101},
  year={2019}
}

@inproceedings{he2024detector,
  title={Detector-free structure from motion},
  author={He, Xingyi and Sun, Jiaming and Wang, Yifan and Peng, Sida and Huang, Qixing and Bao, Hujun and Zhou, Xiaowei},
  booktitle=CVPR,
  pages={21594--21603},
  year={2024}
}

@inproceedings{xu2023iterative,
  title={Iterative geometry encoding volume for stereo matching},
  author={Xu, Gangwei and Wang, Xianqi and Ding, Xiaohuan and Yang, Xin},
  booktitle=CVPR,
  pages={21919--21928},
  year={2023}
}

@article{teed2018deepv2d,
  title={Deepv2d: Video to depth with differentiable structure from motion},
  author={Teed, Zachary and Deng, Jia},
  journal={arXiv preprint arXiv:1812.04605},
  year={2018}
}

@inproceedings{wang2024vggsfm,
  title={Vggsfm: Visual geometry grounded deep structure from motion},
  author={Wang, Jianyuan and Karaev, Nikita and Rupprecht, Christian and Novotny, David},
  booktitle=CVPR,
  pages={21686--21697},
  year={2024}
}

@inproceedings{dust3r,
  title={Dust3r: Geometric 3d vision made easy},
  author={Wang, Shuzhe and Leroy, Vincent and Cabon, Yohann and Chidlovskii, Boris and Revaud, Jerome},
  booktitle=CVPR,
  pages={20697--20709},
  year={2024}
}

@inproceedings{fast3r,
  title={Fast3r: Towards 3d reconstruction of 1000+ images in one forward pass},
  author={Yang, Jianing and Sax, Alexander and Liang, Kevin J and Henaff, Mikael and Tang, Hao and Cao, Ang and Chai, Joyce and Meier, Franziska and Feiszli, Matt},
  booktitle=CVPR,
  pages={21924--21935},
  year={2025}
}

@inproceedings{cut3r,
  title={Continuous 3d perception model with persistent state},
  author={Wang, Qianqian and Zhang, Yifei and Holynski, Aleksander and Efros, Alexei A and Kanazawa, Angjoo},
  booktitle=CVPR,
  pages={10510--10522},
  year={2025}
}

@inproceedings{Mv-dust3r+,
  title={Mv-dust3r+: Single-stage scene reconstruction from sparse views in 2 seconds},
  author={Tang, Zhenggang and Fan, Yuchen and Wang, Dilin and Xu, Hongyu and Ranjan, Rakesh and Schwing, Alexander and Yan, Zhicheng},
  booktitle=CVPR,
  pages={5283--5293},
  year={2025}
}

@inproceedings{must3r,
  title={Must3r: Multi-view network for stereo 3d reconstruction},
  author={Cabon, Yohann and Stoffl, Lucas and Antsfeld, Leonid and Csurka, Gabriela and Chidlovskii, Boris and Revaud, Jerome and Leroy, Vincent},
  booktitle=CVPR,
  pages={1050--1060},
  year={2025}
}

@inproceedings{monst3r,
  title={MonST3R: A Simple Approach for Estimating Geometry in the Presence of Motion},
  author={Zhang, Junyi and Herrmann, Charles and Hur, Junhwa and Jampani, Varun and Darrell, Trevor and Cole, Forrester and Sun, Deqing and Yang, Ming-Hsuan},
  booktitle=ICLR,
  year={2025},
}

@inproceedings{mast3r,
  title={Grounding image matching in 3d with mast3r},
  author={Leroy, Vincent and Cabon, Yohann and Revaud, J{\'e}r{\^o}me},
  booktitle=ECCV,
  pages={71--91},
  year={2024},
  organization={Springer}
}

@inproceedings{flare,
  title={Flare: Feed-forward geometry, appearance and camera estimation from uncalibrated sparse views},
  author={Zhang, Shangzhan and Wang, Jianyuan and Xu, Yinghao and Xue, Nan and Rupprecht, Christian and Zhou, Xiaowei and Shen, Yujun and Wetzstein, Gordon},
  booktitle=CVPR,
  pages={21936--21947},
  year={2025}
}

@article{umeyama2002least,
  title={Least-squares estimation of transformation parameters between two point patterns},
  author={Umeyama, Shinji},
  journal=PAMI,
  volume={13},
  number={4},
  pages={376--380},
  year={2002},
  publisher={IEEE}
}

@inproceedings{ling2024dl3dv,
  title={Dl3dv-10k: A large-scale scene dataset for deep learning-based 3d vision},
  author={Ling, Lu and Sheng, Yichen and Tu, Zhi and Zhao, Wentian and Xin, Cheng and Wan, Kun and Yu, Lantao and Guo, Qianyu and Yu, Zixun and Lu, Yawen and others},
  booktitle=CVPR,
  pages={22160--22169},
  year={2024}
}

@inproceedings{li2018megadepth,
  title={Megadepth: Learning single-view depth prediction from internet photos},
  author={Li, Zhengqi and Snavely, Noah},
  booktitle=CVPR,
  pages={2041--2050},
  year={2018}
}

@article{knapitsch2017tanks,
  title={Tanks and temples: Benchmarking large-scale scene reconstruction},
  author={Knapitsch, Arno and Park, Jaesik and Zhou, Qian-Yi and Koltun, Vladlen},
  journal=TOG,
  volume={36},
  number={4},
  pages={1--13},
  year={2017},
  publisher={ACM New York, NY, USA}
}

@inproceedings{silberman2012indoor,
  title={Indoor segmentation and support inference from rgbd images},
  author={Silberman, Nathan and Hoiem, Derek and Kohli, Pushmeet and Fergus, Rob},
  booktitle=ECCV,
  pages={746--760},
  year={2012},
  organization={Springer}
}

@article{geiger2013vision,
  title={Vision meets robotics: The kitti dataset},
  author={Geiger, Andreas and Lenz, Philip and Stiller, Christoph and Urtasun, Raquel},
  journal={The international journal of robotics research},
  volume={32},
  number={11},
  pages={1231--1237},
  year={2013},
  publisher={Sage Publications Sage UK: London, England}
}

@inproceedings{wang2025moge,
  title={Moge: Unlocking accurate monocular geometry estimation for open-domain images with optimal training supervision},
  author={Wang, Ruicheng and Xu, Sicheng and Dai, Cassie and Xiang, Jianfeng and Deng, Yu and Tong, Xin and Yang, Jiaolong},
  booktitle=CVPR,
  pages={5261--5271},
  year={2025}
}

@inproceedings{wang2023posediffusion,
  title={Posediffusion: Solving pose estimation via diffusion-aided bundle adjustment},
  author={Wang, Jianyuan and Rupprecht, Christian and Novotny, David},
  booktitle=CVPR,
  pages={9773--9783},
  year={2023}
}

@inproceedings{zhang2023geomvsnet,
  title={Geomvsnet: Learning multi-view stereo with geometry perception},
  author={Zhang, Zhe and Peng, Rui and Hu, Yuxi and Wang, Ronggang},
  booktitle=CVPR,
  pages={21508--21518},
  year={2023}
}

@article{BiRefNet,
  title={Bilateral Reference for High-Resolution Dichotomous Image Segmentation},
  author={Zheng, Peng and Gao, Dehong and Fan, Deng-Ping and Liu, Li and Laaksonen, Jorma and Ouyang, Wanli and Sebe, Nicu},
  journal={CAAI Artificial Intelligence Research},
  year={2024}
}

@inproceedings{zhu2024nicerslam,
  title={Nicer-slam: Neural implicit scene encoding for rgb slam},
  author={Zhu, Zihan and Peng, Songyou and Larsson, Viktor and Cui, Zhaopeng and Oswald, Martin R and Geiger, Andreas and Pollefeys, Marc},
  booktitle={3DV},
  pages={42--52},
  year={2024},
  organization={IEEE}
}

@misc{keetha2025mapanything,
  title={{MapAnything}: Universal Feed-Forward Metric {3D} Reconstruction},
  author={Nikhil Keetha and Norman M\"{u}ller and Johannes Sch\"{o}nberger and Lorenzo Porzi and Yuchen Zhang and Tobias Fischer and Arno Knapitsch and Duncan Zauss and Ethan Weber and Nelson Antunes and Jonathon Luiten and Manuel Lopez-Antequera and Samuel Rota Bul\`{o} and Christian Richardt and Deva Ramanan and Sebastian Scherer and Peter Kontschieder},
  note={arXiv preprint arXiv:2509.13414},
  year={2025}
}

@misc{wang2025pi3,
      title={$\pi^3$: Scalable Permutation-Equivariant Visual Geometry Learning}, 
      author={Yifan Wang and Jianjun Zhou and Haoyi Zhu and Wenzheng Chang and Yang Zhou and Zizun Li and Junyi Chen and Jiangmiao Pang and Chunhua Shen and Tong He},
      year={2025},
      eprint={2507.13347},
      archivePrefix={arXiv},
      primaryClass={cs.CV},
      url={https://arxiv.org/abs/2507.13347}, 
}

@inproceedings{guo2025multi,
  title={Multi-view reconstruction via sfm-guided monocular depth estimation},
  author={Guo, Haoyu and Zhu, He and Peng, Sida and Lin, Haotong and Yan, Yunzhi and Xie, Tao and Wang, Wenguan and Zhou, Xiaowei and Bao, Hujun},
  booktitle=CVPR,
  pages={5272--5282},
  year={2025}
}

@inproceedings{pan2024global,
  title={Global structure-from-motion revisited},
  author={Pan, Linfei and Bar{\'a}th, D{\'a}niel and Pollefeys, Marc and Sch{\"o}nberger, Johannes L},
  booktitle=ECCV,
  pages={58--77},
  year={2024},
  organization={Springer}
}

@inproceedings{murai2025mast3r,
  title={MASt3R-SLAM: Real-time dense SLAM with 3D reconstruction priors},
  author={Murai, Riku and Dexheimer, Eric and Davison, Andrew J},
  booktitle=CVPR,
  pages={16695--16705},
  year={2025}
}

@inproceedings{jang2025pow3r,
  title={Pow3r: Empowering unconstrained 3d reconstruction with camera and scene priors},
  author={Jang, Wonbong and Weinzaepfel, Philippe and Leroy, Vincent and Agapito, Lourdes and Revaud, Jerome},
  booktitle=CVPR,
  pages={1071--1081},
  year={2025}
}

@article{bochkovskii2024depth,
  title={Depth pro: Sharp monocular metric depth in less than a second},
  author={Bochkovskii, Aleksei and Delaunoy, Ama{\~A}{\c{G}}l and Germain, Hugo and Santos, Marcel and Zhou, Yichao and Richter, Stephan R and Koltun, Vladlen},
  journal={arXiv preprint arXiv:2410.02073},
  year={2024}
}

@inproceedings{yin2023metric3d,
  title={Metric3d: Towards zero-shot metric 3d prediction from a single image},
  author={Yin, Wei and Zhang, Chi and Chen, Hao and Cai, Zhipeng and Yu, Gang and Wang, Kaixuan and Chen, Xiaozhi and Shen, Chunhua},
  booktitle={CVPR},
  pages={9043--9053},
  year={2023}
}

@article{hu2024metric3dv2,
  title={Metric3D v2: A Versatile Monocular Geometric Foundation Model for Zero-shot Metric Depth and Surface Normal Estimation},
  author={Hu, Mu and Yin, Wei and Zhang, Chi and Cai, Zhipeng and Long, Xiaoxiao and Chen, Hao and Wang, Kaixuan and Yu, Gang and Shen, Chunhua and Shen, Shaojie},
  journal={TPAMI},
  year={2024}
}

@inproceedings{ke2024repurposing,
  title={Repurposing diffusion-based image generators for monocular depth estimation},
  author={Ke, Bingxin and Obukhov, Anton and Huang, Shengyu and Metzger, Nando and Daudt, Rodrigo Caye and Schindler, Konrad},
  booktitle={CVPR},
  pages={9492--9502},
  year={2024}
}

@inproceedings{peebles2023scalable,
  title={Scalable diffusion models with transformers},
  author={Peebles, William and Xie, Saining},
  booktitle={Proceedings of the IEEE/CVF international conference on computer vision},
  pages={4195--4205},
  year={2023}
}

@article{abdel2015direct,
  title={Direct linear transformation from comparator coordinates into object space coordinates in close-range photogrammetry},
  author={Abdel-Aziz, Yousset I and Karara, Hauck Michael and Hauck, Michael},
  journal={Photogrammetric engineering \& remote sensing},
  volume={81},
  number={2},
  pages={103--107},
  year={2015},
  publisher={Elsevier}
}

@inproceedings{pan2023aria,
  title={Aria digital twin: A new benchmark dataset for egocentric 3d machine perception},
  author={Pan, Xiaqing and Charron, Nicholas and Yang, Yongqian and Peters, Scott and Whelan, Thomas and Kong, Chen and Parkhi, Omkar and Newcombe, Richard and Ren, Yuheng Carl},
  booktitle={Proceedings of the IEEE/CVF International Conference on Computer Vision},
  pages={20133--20143},
  year={2023}
}

@inproceedings{yao2020blendedmvs,
  title={Blendedmvs: A large-scale dataset for generalized multi-view stereo networks},
  author={Yao, Yao and Luo, Zixin and Li, Shiwei and Zhang, Jingyang and Ren, Yufan and Zhou, Lei and Fang, Tian and Quan, Long},
  booktitle={Proceedings of the IEEE/CVF conference on computer vision and pattern recognition},
  pages={1790--1799},
  year={2020}
}

@inproceedings{roberts2021hypersim,
  title={Hypersim: A photorealistic synthetic dataset for holistic indoor scene understanding},
  author={Roberts, Mike and Ramapuram, Jason and Ranjan, Anurag and Kumar, Atulit and Bautista, Miguel Angel and Paczan, Nathan and Webb, Russ and Susskind, Joshua M},
  booktitle={Proceedings of the IEEE/CVF international conference on computer vision},
  pages={10912--10922},
  year={2021}
}

@inproceedings{arnold2022map,
  title={Map-free visual relocalization: Metric pose relative to a single image},
  author={Arnold, Eduardo and Wynn, Jamie and Vicente, Sara and Garcia-Hernando, Guillermo and Monszpart, Aron and Prisacariu, Victor and Turmukhambetov, Daniyar and Brachmann, Eric},
  booktitle={European Conference on Computer Vision},
  pages={690--708},
  year={2022},
  organization={Springer}
}

@inproceedings{jiang2025megasynth,
  title={Megasynth: Scaling up 3d scene reconstruction with synthesized data},
  author={Jiang, Hanwen and Xu, Zexiang and Xie, Desai and Chen, Ziwen and Jin, Haian and Luan, Fujun and Shu, Zhixin and Zhang, Kai and Bi, Sai and Sun, Xin and others},
  booktitle={Proceedings of the Computer Vision and Pattern Recognition Conference},
  pages={16441--16452},
  year={2025}
}

@inproceedings{DeepMVS,
      author       = "Huang, Po-Han and Matzen, Kevin and Kopf, Johannes and Ahuja, Narendra and Huang, Jia-Bin",
      title        = "DeepMVS: Learning Multi-View Stereopsis",
      booktitle    = "IEEE Conference on Computer Vision and Pattern Recognition (CVPR)",
      year         = "2018"
    }

@inproceedings{deitke2023objaverse,
  title={Objaverse: A universe of annotated 3d objects},
  author={Deitke, Matt and Schwenk, Dustin and Salvador, Jordi and Weihs, Luca and Michel, Oscar and VanderBilt, Eli and Schmidt, Ludwig and Ehsani, Kiana and Kembhavi, Aniruddha and Farhadi, Ali},
  booktitle={Proceedings of the IEEE/CVF conference on computer vision and pattern recognition},
  pages={13142--13153},
  year={2023}
}

@article{xiang2024structured,
    title   = {Structured 3D Latents for Scalable and Versatile 3D Generation},
    author  = {Xiang, Jianfeng and Lv, Zelong and Xu, Sicheng and Deng, Yu and Wang, Ruicheng and Zhang, Bowen and Chen, Dong and Tong, Xin and Yang, Jiaolong},
    journal = {arXiv preprint arXiv:2412.01506},
    year    = {2024}
}

@inproceedings{zheng2023pointodyssey,
  title={Pointodyssey: A large-scale synthetic dataset for long-term point tracking},
  author={Zheng, Yang and Harley, Adam W and Shen, Bokui and Wetzstein, Gordon and Guibas, Leonidas J},
  booktitle={Proceedings of the IEEE/CVF International Conference on Computer Vision},
  pages={19855--19865},
  year={2023}
}

@article{straub2019replica,
  title={The replica dataset: A digital replica of indoor spaces},
  author={Straub, Julian and Whelan, Thomas and Ma, Lingni and Chen, Yufan and Wijmans, Erik and Green, Simon and Engel, Jakob J and Mur-Artal, Raul and Ren, Carl and Verma, Shobhit and others},
  journal={arXiv preprint arXiv:1906.05797},
  year={2019}
}

@inproceedings{mccormac2017scenenet,
  title={Scenenet rgb-d: Can 5m synthetic images beat generic imagenet pre-training on indoor segmentation?},
  author={McCormac, John and Handa, Ankur and Leutenegger, Stefan and Davison, Andrew J},
  booktitle={Proceedings of the IEEE International Conference on Computer Vision},
  pages={2678--2687},
  year={2017}
}

@inproceedings{tartanair2020iros,
  title={Tartanair: A dataset to push the limits of visual slam},
  author={Wang, Wenshan and Zhu, Delong and Wang, Xiangwei and Hu, Yaoyu and Qiu, Yuheng and Wang, Chen and Hu, Yafei and Kapoor, Ashish and Scherer, Sebastian},
  booktitle={2020 IEEE/RSJ International Conference on Intelligent Robots and Systems (IROS)},
  pages={4909--4916},
  year={2020},
  organization={IEEE}
}

@misc{cabon2020vkitti2,
  title={Virtual KITTI 2},
  author={Cabon, Yohann and Murray, Naila and Humenberger, Martin},
  year={2020},
  eprint={2001.10773},
  archivePrefix={arXiv},
  primaryClass={cs.CV}
}

@misc{xia2024rgbd,
        title={RGBD Objects in the Wild: Scaling Real-World 3D Object Learning from RGB-D Videos}, 
        author={Hongchi Xia and Yang Fu and Sifei Liu and Xiaolong Wang},
        year={2024},
        eprint={2401.12592},
        archivePrefix={arXiv},
        primaryClass={cs.CV}
      }

@inproceedings{wu2023omniobject3d,
    author = {Tong Wu and Jiarui Zhang and Xiao Fu and Yuxin Wang and Jiawei Ren, 
    Liang Pan and Wayne Wu and Lei Yang and Jiaqi Wang and Chen Qian and Dahua Lin and Ziwei Liu},
    title = {OmniObject3D: Large-Vocabulary 3D Object Dataset for Realistic Perception, 
    Reconstruction and Generation},
    booktitle={IEEE/CVF Conference on Computer Vision and Pattern Recognition (CVPR)},
    year={2023}
}

@article{wang2019irs,
  title={Irs: A large naturalistic indoor robotics stereo dataset to train deep models for disparity and surface normal estimation},
  author={Wang, Qiang and Zheng, Shizhen and Yan, Qingsong and Deng, Fei and Zhao, Kaiyong and Chu, Xiaowen},
  journal={arXiv preprint arXiv:1912.09678},
  year={2019}
}

@inproceedings{mehl2023spring,
  title={Spring: A high-resolution high-detail dataset and benchmark for scene flow, optical flow and stereo},
  author={Mehl, Lukas and Schmalfuss, Jenny and Jahedi, Azin and Nalivayko, Yaroslava and Bruhn, Andr{\'e}s},
  booktitle={Proceedings of the IEEE/CVF Conference on Computer Vision and Pattern Recognition},
  pages={4981--4991},
  year={2023}
}

@inproceedings{zhang2018unrealstereo,
  title={Unrealstereo: Controlling hazardous factors to analyze stereo vision},
  author={Zhang, Yi and Qiu, Weichao and Chen, Qi and Hu, Xiaolin and Yuille, Alan},
  booktitle={2018 International Conference on 3D Vision (3DV)},
  pages={228--237},
  year={2018},
  organization={IEEE}
}

@article{wang2020flow,
  title={Flow-motion and depth network for monocular stereo and beyond},
  author={Wang, Kaixuan and Shen, Shaojie},
  journal={IEEE Robotics and Automation Letters},
  volume={5},
  number={2},
  pages={3307--3314},
  year={2020},
  publisher={IEEE}
}

@article{gil2021online,
  title={Online training of stereo self-calibration using monocular depth estimation},
  author={Gil, Yotam and Elmalem, Shay and Haim, Harel and Marom, Emanuel and Giryes, Raja},
  journal={IEEE Transactions on Computational Imaging},
  volume={7},
  pages={812--823},
  year={2021},
  publisher={IEEE}
}

@article{niklaus20193d,
  title={3d ken burns effect from a single image},
  author={Niklaus, Simon and Mai, Long and Yang, Jimei and Liu, Feng},
  journal={ACM Transactions on Graphics (ToG)},
  volume={38},
  number={6},
  pages={1--15},
  year={2019},
  publisher={ACM New York, NY, USA}
}

@inproceedings{li2023matrixcity,
  title={Matrixcity: A large-scale city dataset for city-scale neural rendering and beyond},
  author={Li, Yixuan and Jiang, Lihan and Xu, Linning and Xiangli, Yuanbo and Wang, Zhenzhi and Lin, Dahua and Dai, Bo},
  booktitle={Proceedings of the IEEE/CVF International Conference on Computer Vision},
  pages={3205--3215},
  year={2023}
}

@inproceedings{le2021eden,
  title={Eden: Multimodal synthetic dataset of enclosed garden scenes},
  author={Le, Hoang-An and Mensink, Thomas and Das, Partha and Karaoglu, Sezer and Gevers, Theo},
  booktitle={Proceedings of the IEEE/CVF Winter Conference on Applications of Computer Vision},
  pages={1579--1589},
  year={2021}
}

@article{replica19arxiv,
  title =   {The {R}eplica Dataset: A Digital Replica of Indoor Spaces},
  author =  {Julian Straub and Thomas Whelan and Lingni Ma and Yufan Chen and Erik Wijmans and Simon Green and Jakob J. Engel and Raul Mur-Artal and Carl Ren and Shobhit Verma and Anton Clarkson and Mingfei Yan and Brian Budge and Yajie Yan and Xiaqing Pan and June Yon and Yuyang Zou and Kimberly Leon and Nigel Carter and Jesus Briales and  Tyler Gillingham and  Elias Mueggler and Luis Pesqueira and Manolis Savva and Dhruv Batra and Hauke M. Strasdat and Renzo De Nardi and Michael Goesele and Steven Lovegrove and Richard Newcombe },
  journal = {arXiv preprint arXiv:1906.05797},
  year =    {2019}
}

@inproceedings{zheng2020structured3d,
  title={Structured3d: A large photo-realistic dataset for structured 3d modeling},
  author={Zheng, Jia and Zhang, Junfei and Li, Jing and Tang, Rui and Gao, Shenghua and Zhou, Zihan},
  booktitle={European Conference on Computer Vision},
  pages={519--535},
  year={2020},
  organization={Springer}
}

@article{GOMEZ2025130038, title = {All for one, and one for all: UrbanSyn Dataset, the third Musketeer of synthetic driving scenes},
journal = {Neurocomputing},
volume = {637},
pages = {130038},
year = {2025},
issn = {0925-2312},
doi = {https://doi.org/10.1016/j.neucom.2025.130038},
url = {https://www.sciencedirect.com/science/article/pii/S0925231225007106},
author = {Jose L. Gómez and Manuel Silva and Antonio Seoane and Agnés Borràs and Mario Noriega and German Ros and Jose A. Iglesias-Guitian and Antonio M. López},
}

@inproceedings{chen2024mvsplat,
  title={Mvsplat: Efficient 3d gaussian splatting from sparse multi-view images},
  author={Chen, Yuedong and Xu, Haofei and Zheng, Chuanxia and Zhuang, Bohan and Pollefeys, Marc and Geiger, Andreas and Cham, Tat-Jen and Cai, Jianfei},
  booktitle=ECCV,
  pages={370--386},
  year={2024},
  organization={Springer}
}

@inproceedings{charatan2024pixelsplat,
  title={pixelsplat: 3d gaussian splats from image pairs for scalable generalizable 3d reconstruction},
  author={Charatan, David and Li, Sizhe Lester and Tagliasacchi, Andrea and Sitzmann, Vincent},
  booktitle=CVPR,
  pages={19457--19467},
  year={2024}
}

@misc{mtsdrones2024,
  author = {MTS Drones},
  title = {Drone Australia Gliding Ep025: Sydney Views | Opera House, Harbour Bridge \& Hyde Park | DJI Mavic 4K},
  year = {2024},
  howpublished = {\url{https://www.youtube.com/watch?v=qbgKDaGraTA}},
  note = {Accessed: Sep. 25, 2025. Used under YouTube Standard License.}
}

@inproceedings{zamir2018taskonomy,
  title={Taskonomy: Disentangling Task Transfer Learning},
  author={Zamir, Amir R and Sax, Alexander and and Shen, William B and Guibas, Leonidas and Malik, Jitendra and Savarese, Silvio},
  booktitle=cvpr,
  year={2018},
  organization={IEEE}
}

@inproceedings{ddad,
  author = {Vitor Guizilini and Rares Ambrus and Sudeep Pillai and Allan Raventos and Adrien Gaidon},
  title = {3D Packing for Self-Supervised Monocular Depth Estimation},
  booktitle = cvpr,
  primaryClass = {cs.CV},
  year = {2020},
}

@inproceedings{wilson2023argoverse,
  title={Argoverse 2: Next generation datasets for self-driving perception and forecasting},
  author={Wilson, Benjamin and Qi, William and Agarwal, Tanmay and Lambert, John and Singh, Jagjeet and Khandelwal, Siddhesh and Pan, Bowen and Kumar, Ratnesh and Hartnett, Andrew and Pontes, Jhony Kaesemodel and others},
  booktitle = nips,
  year = 2021,
}

@inproceedings{xiao2021pandaset,
  title={Pandaset: Advanced sensor suite dataset for autonomous driving},
  author={Xiao, Pengchuan and Shao, Zhenlei and Hao, Steven and Zhang, Zishuo and Chai, Xiaolin and Jiao, Judy and Li, Zesong and Wu, Jian and Sun, Kai and Jiang, Kun and others},
  booktitle={2021 IEEE international intelligent transportation systems conference (ITSC)},
  pages={3095--3101},
  year={2021},
  organization={IEEE}
}

@InProceedings{Sun_2020_CVPR, author = {Sun, Pei and Kretzschmar, Henrik and Dotiwalla, Xerxes and Chouard, Aurelien and Patnaik, Vijaysai and Tsui, Paul and Guo, James and Zhou, Yin and Chai, Yuning and Caine, Benjamin and Vasudevan, Vijay and Han, Wei and Ngiam, Jiquan and Zhao, Hang and Timofeev, Aleksei and Ettinger, Scott and Krivokon, Maxim and Gao, Amy and Joshi, Aditya and Zhang, Yu and Shlens, Jonathon and Chen, Zhifeng and Anguelov, Dragomir}, title = {Scalability in Perception for Autonomous Driving: Waymo Open Dataset}, booktitle = {Proceedings of the IEEE/CVF Conference on Computer Vision and Pattern Recognition (CVPR)}, month = {June}, year = {2020} }

@article{Gehrig21ral,
  title={Dsec: A stereo event camera dataset for driving scenarios},
  author={Gehrig, Mathias and Aarents, Willem and Gehrig, Daniel and Scaramuzza, Davide},
  journal={IEEE Robotics and Automation Letters},
  volume={6},
  number={3},
  pages={4947--4954},
  year={2021},
  publisher={IEEE}
}

@inproceedings{yang2019drivingstereo,
    title={DrivingStereo: A Large-Scale Dataset for Stereo Matching in Autonomous Driving Scenarios},
    author={Yang, Guorun and Song, Xiao and Huang, Chaoqin and Deng, Zhidong and Shi, Jianping and Zhou, Bolei},
    booktitle={IEEE Conference on Computer Vision and Pattern Recognition (CVPR)},
    year={2019}
}

@inproceedings{cordts2016cityscapes,
  title={The cityscapes dataset for semantic urban scene understanding},
  author={Cordts, Marius and Omran, Mohamed and Ramos, Sebastian and Rehfeld, Timo and Enzweiler, Markus and Benenson, Rodrigo and Franke, Uwe and Roth, Stefan and Schiele, Bernt},
  booktitle={Proceedings of the IEEE conference on computer vision and pattern recognition},
  pages={3213--3223},
  year={2016}
}

@inproceedings{wen2025foundationstereo,
  title={Foundationstereo: Zero-shot stereo matching},
  author={Wen, Bowen and Trepte, Matthew and Aribido, Joseph and Kautz, Jan and Gallo, Orazio and Birchfield, Stan},
  booktitle=cvpr,
  pages={5249--5260},
  year={2025}
}

@article{cho2021diml,
  title={Diml/cvl rgb-d dataset: 2m rgb-d images of natural indoor and outdoor scenes},
  author={Cho, Jaehoon and Min, Dongbo and Kim, Youngjung and Sohn, Kwanghoon},
  journal={arXiv preprint arXiv:2110.11590},
  year={2021}
}

@inproceedings{piccinelli2024unidepth,
  title={UniDepth: Universal monocular metric depth estimation},
  author={Piccinelli, Luigi and Yang, Yung-Hsu and Sakaridis, Christos and Segu, Mattia and Li, Siyuan and Van Gool, Luc and Yu, Fisher},
  booktitle=cvpr,
  pages={10106--10116},
  year={2024}
}

@article{piccinelli2025unidepthv2,
  title={Unidepthv2: Universal monocular metric depth estimation made simpler},
  author={Piccinelli, Luigi and Sakaridis, Christos and Yang, Yung-Hsu and Segu, Mattia and Li, Siyuan and Abbeloos, Wim and Van Gool, Luc},
  journal={arXiv preprint arXiv:2502.20110},
  year={2025}
}

@article{hu2024metric3d,
  title={Metric3d v2: A versatile monocular geometric foundation model for zero-shot metric depth and surface normal estimation},
  author={Hu, Mu and Yin, Wei and Zhang, Chi and Cai, Zhipeng and Long, Xiaoxiao and Chen, Hao and Wang, Kaixuan and Yu, Gang and Shen, Chunhua and Shen, Shaojie},
  journal={IEEE Transactions on Pattern Analysis and Machine Intelligence},
  year={2024},
  publisher={IEEE}
}

@inproceedings{bochkovskiydepth,
  title={Depth Pro: Sharp Monocular Metric Depth in Less Than a Second},
  author={Bochkovskiy, Alexey and Delaunoy, Ama{\"e}l and Germain, Hugo and Santos, Marcel and Zhou, Yichao and Richter, Stephan and Koltun, Vladlen},
  booktitle=iclr,
  year={2025}
}

@inproceedings{ye2024no,
  title={No pose, no problem: Surprisingly simple 3d gaussian splats from sparse unposed images},
  author={Ye, Botao and Liu, Sifei and Xu, Haofei and Li, Xueting and Pollefeys, Marc and Yang, Ming-Hsuan and Peng, Songyou},
  booktitle=iclr,
  year={2024}
}

@article{smart2024splatt3r,
  title={Splatt3r: Zero-shot gaussian splatting from uncalibrated image pairs},
  author={Smart, Brandon and Zheng, Chuanxia and Laina, Iro and Prisacariu, Victor Adrian},
  journal={arXiv preprint arXiv:2408.13912},
  year={2024}
}

@article{jiang2025anysplat,
  title={AnySplat: Feed-forward 3D Gaussian Splatting from Unconstrained Views},
  author={Jiang, Lihan and Mao, Yucheng and Xu, Linning and Lu, Tao and Ren, Kerui and Jin, Yichen and Xu, Xudong and Yu, Mulin and Pang, Jiangmiao and Zhao, Feng and others},
  journal=TOG,
  year={2025}
}

@inproceedings{guedon2024sugar,
  title={Sugar: Surface-aligned gaussian splatting for efficient 3d mesh reconstruction and high-quality mesh rendering},
  author={Gu{\'e}don, Antoine and Lepetit, Vincent},
  booktitle=cvpr,
  pages={5354--5363},
  year={2024}
}

@article{deng2025sailrecon,
  title={SAIL-Recon: Large SfM by Augmenting Scene Regression with Localization},
  author={Deng, Junyuan and Li, Heng and Xie, Tao and Ren, Weiqiang and Zhang, Qian and Tan, Ping and Guo, Xiaoyang},
  journal={arXiv preprint arXiv:2508.17972},
  year={2025}
}

@article{maggio2025vggtslam,
  title={VGGT-SLAM: Dense RGB SLAM Optimized on the SL(4) Manifold},
  author={Maggio, Dominic and Lim, Hyungtae and Carlone, Luca},
  journal={arXiv preprint arXiv:2505.12549},
  year={2025}
}

@article{deng2025vggtlong,
  title={VGGT-Long: Chunk it, Loop it, Align it -- Pushing VGGT's Limits on Kilometer-scale Long RGB Sequences},
  author={Deng, Kai and Ti, Zexin and Xu, Jiawei and Yang, Jian and Xie, Jin},
  journal={arXiv preprint arXiv:2507.11539},
  year={2025}
}

@article{sitzmann2019scene,
  title={Scene representation networks: Continuous 3d-structure-aware neural scene representations},
  author={Sitzmann, Vincent and Zollh{\"o}fer, Michael and Wetzstein, Gordon},
  journal=NIPS,
  volume={32},
  year={2019}
}

@article{sitzmann2021light,
  title={Light field networks: Neural scene representations with single-evaluation rendering},
  author={Sitzmann, Vincent and Rezchikov, Semon and Freeman, Bill and Tenenbaum, Josh and Durand, Fredo},
  journal=NIPS,
  volume={34},
  pages={19313--19325},
  year={2021}
}

@inproceedings{govindarajan2025radiant,
  title={Radiant foam: Real-time differentiable ray tracing},
  author={Govindarajan, Shrisudhan and Rebain, Daniel and Yi, Kwang Moo and Tagliasacchi, Andrea},
  booktitle=ICCV,
  year={2025}
}

@article{kerbl20233d,
  title={3D Gaussian splatting for real-time radiance field rendering.},
  author={Kerbl, Bernhard and Kopanas, Georgios and Leimk{\"u}hler, Thomas and Drettakis, George},
  journal=TOG,
  volume={42},
  number={4},
  pages={139--1},
  year={2023}
}

@inproceedings{yu2021pixelnerf,
  title={pixelnerf: Neural radiance fields from one or few images},
  author={Yu, Alex and Ye, Vickie and Tancik, Matthew and Kanazawa, Angjoo},
  booktitle=CVPR,
  pages={4578--4587},
  year={2021}
}

@inproceedings{chen2021mvsnerf,
  title={Mvsnerf: Fast generalizable radiance field reconstruction from multi-view stereo},
  author={Chen, Anpei and Xu, Zexiang and Zhao, Fuqiang and Zhang, Xiaoshuai and Xiang, Fanbo and Yu, Jingyi and Su, Hao},
  booktitle=CVPR,
  pages={14124--14133},
  year={2021}
}

@inproceedings{xu2024murf,
  title={MuRF: multi-baseline radiance fields},
  author={Xu, Haofei and Chen, Anpei and Chen, Yuedong and Sakaridis, Christos and Zhang, Yulun and Pollefeys, Marc and Geiger, Andreas and Yu, Fisher},
  booktitle=CVPR,
  pages={20041--20050},
  year={2024}
}

@inproceedings{lin2022efficient,
  title={Efficient neural radiance fields for interactive free-viewpoint video},
  author={Lin, Haotong and Peng, Sida and Xu, Zhen and Yan, Yunzhi and Shuai, Qing and Bao, Hujun and Zhou, Xiaowei},
  booktitle={SIGGRAPH Asia 2022 Conference Papers},
  pages={1--9},
  year={2022}
}

@article{chen2025explicit,
  title={Explicit correspondence matching for generalizable neural radiance fields},
  author={Chen, Yuedong and Xu, Haofei and Wu, Qianyi and Zheng, Chuanxia and Cham, Tat-Jen and Cai, Jianfei},
  journal=PAMI,
  year={2025},
  publisher={IEEE}
}

@inproceedings{buehler2001unstructured,
  title={Unstructured lumigraph rendering},
  author={Buehler, Chris and Bosse, Michael and McMillan, Leonard and Gortler, Steven and Cohen, Michael},
  booktitle={Proceedings of the 28th annual conference on Computer graphics and interactive techniques},
  pages={425--432},
  year={2001}
}

@inproceedings{heigl1999plenoptic,
  title={Plenoptic modeling and rendering from image sequences taken by a hand-held camera},
  author={Heigl, Benno and Koch, Reinhard and Pollefeys, Marc and Denzler, Joachim and Van Gool, Luc},
  booktitle={Mustererkennung 1999: 21. DAGM-Symposium Bonn, 15.--17. September 1999},
  pages={94--101},
  year={1999},
  organization={Springer}
}

@article{levoy1996light,
  title={Light field rendering},
  author={Levoy, Marc and Hanrahan, Pat},
  journal=TOG,
  year={1996}
}

@inproceedings{hong2024lrm,
  title={Lrm: Large reconstruction model for single image to 3d},
  author={Hong, Yicong and Zhang, Kai and Gu, Jiuxiang and Bi, Sai and Zhou, Yang and Liu, Difan and Liu, Feng and Sunkavalli, Kalyan and Bui, Trung and Tan, Hao},
  booktitle=ICLR,
  year={2024}
}

@misc{SVLightVerse2025,
  title        = {SVLightVerse: Large-scale Photorealistic Indoor Dataset with Spatially-Varying HDRI Lighting},
  author       = {Zhu, Rui and Second Author and Third Author},
  year         = {2025},
  howpublished = {\url{https://jerrypiglet.github.io/SVLightVerse/}},
  note         = {Project page. Affiliations: University of California San Diego; PICO (ByteDance); KooLab (Manycore); Rembrand. Accessed: 2025-11-11}
}

@misc{zhou2025omniworld,
      title={OmniWorld: A Multi-Domain and Multi-Modal Dataset for 4D World Modeling}, 
      author={Yang Zhou and Yifan Wang and Jianjun Zhou and Wenzheng Chang and Haoyu Guo and Zizun Li and Kaijing Ma and Xinyue Li and Yating Wang and Haoyi Zhu and Mingyu Liu and Dingning Liu and Jiange Yang and Zhoujie Fu and Junyi Chen and Chunhua Shen and Jiangmiao Pang and Kaipeng Zhang and Tong He},
      year={2025},
      eprint={2509.12201},
      archivePrefix={arXiv},
      primaryClass={cs.CV},
      url={https://arxiv.org/abs/2509.12201}, 
}

\clearpage

\beginappendix

\appendix

While synthetic datasets provide large-scale training data with ground-truth depth annotations, many contain quality issues such as invalid backgrounds, spatial misalignments, clipping artifacts, and erroneous depth values that can degrade model training. We therefore apply careful preprocessing to filter problematic samples and clip unrealistic depth ranges, ensuring high-quality supervision for our teacher model.
\section{Data Processing}

We preprocess the following raw training datasets as follows to train the teacher model. 

\paragraph{TartanAir.} We remove the {\tt amusement} scene from training due to its invalid background (skybox) (Fig.~\ref{fig:tartanair_problem}). We clip the maximum depth values of {\tt carwelding}, {\tt hospital}, {\tt ocean}, {\tt office} and {\tt office2} scenes at 80, 30, 1000, 30 and 30, respectively. 

\begin{figure}[h]
    \centering
    \includegraphics[width=1\linewidth]{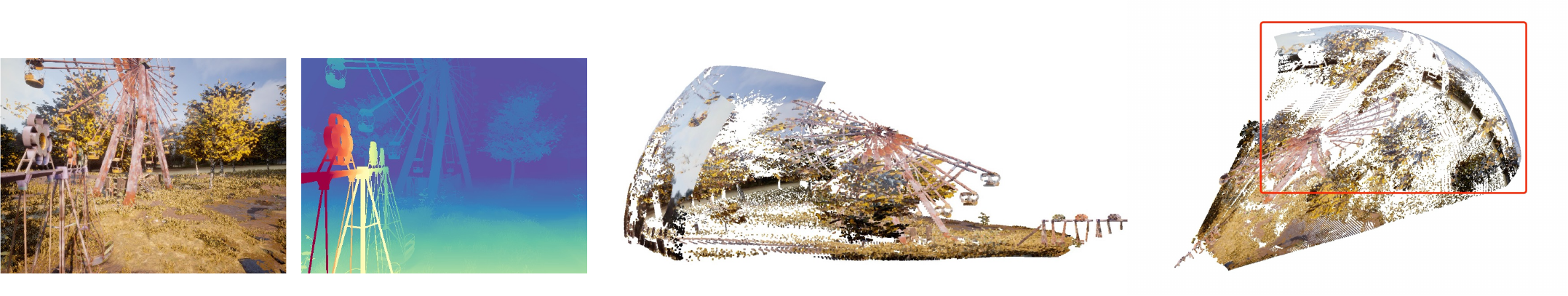}
    \caption{\textbf{Invalid background of {\tt amusement} scene in TartanAir dataset.}}
    \label{fig:tartanair_problem}
\end{figure}

\paragraph{IRS.} We noticed that some of the scenes in IRS exhibit spatial misalignment between the image and depth maps (Fig.~\ref{fig:irs_problem}). To filter those samples with image-depth misalignment, we first run Canny edge detectors on both the image (converted to grayscale) and depth map to extract the boundaries. Next, we dilate the boundaries by 1 pixel and compute the percentage of intersection between image and depth boundaries. Then, we further dilate the boundaries by 3 pixels and compute the intersection between image and depth boundaries. Finally, we compute the ratio between the two intersections and remove samples whose ratio falls below a certain threshold. 

\begin{figure}[h]
    \centering
    \includegraphics[width=1.0\linewidth]{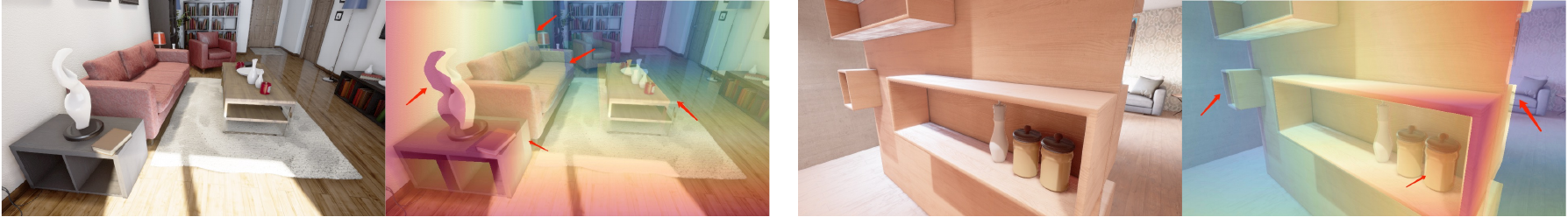}
    \caption{\textbf{Image-depth spatial misalignment in IRS dataset.}}
    \label{fig:irs_problem}
\end{figure}

\paragraph{UnrealStereo4K.} We remove the scene {\tt 00003} which consists of large erroneous regions. We also remove the scene {\tt 00004} which suffers from clipping issue. For scene {\tt 00008}, we remove samples where the sea does not have depth values (images 9-13, 23-29, 80-82, 86-88, 96, 103-111, 126-136, 144-145, 148-154, 173-176, 178-179, 186-187, 191-192, 198-199) (Fig.~\ref{fig:unrealstereo4k_problem}).

\begin{figure}[h]
    \centering
    \includegraphics[width=1.0\linewidth]{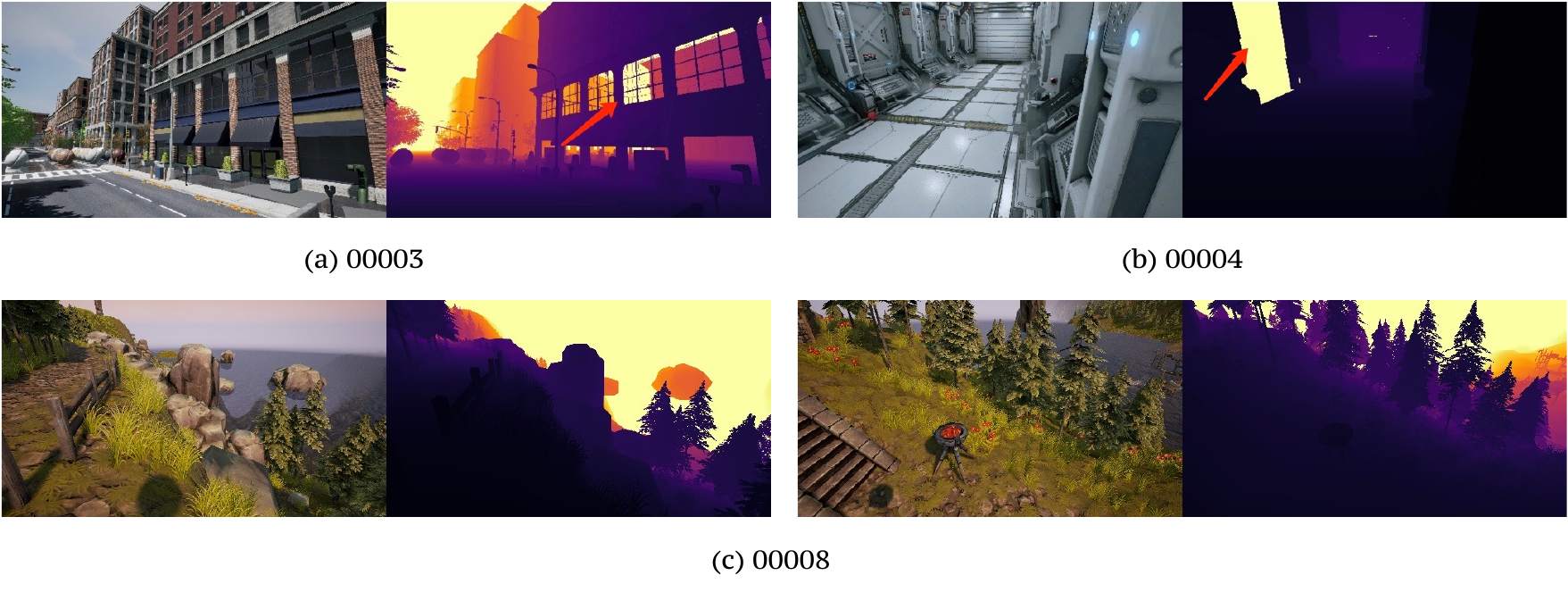}
    \caption{\textbf{Erroneous samples in UnrealStereo4K dataset.} The windows in scene {\tt 00003} are transparent; scene {\tt 00004} suffers from clipping issue; the sea in scene {\tt 00008} does not have depth values.}
    \label{fig:unrealstereo4k_problem}
\end{figure}

\paragraph{GTA-SfM.} We clip the maximum depth value at 1000.

\paragraph{Kenburns.} We clip the depth value at 50,000 following GitHub issue~\footnote{\url{https://github.com/sniklaus/3d-ken-burns/issues/40}}.

\paragraph{PointOdyssey.} We remove the two scenes {\tt animal2\_s} and {\tt dancingroom3\_3rd} where the ground depth is incorrect~\footnote{\url{https://github.com/y-zheng18/point_odyssey/issues/6}}.

\paragraph{TRELLIS.} We remove all samples with suffix {\tt stl}, {\tt abc}, {\tt STL}, {\tt PLY}, {\tt ply} as we noticed that these samples do not contain texture.

\paragraph{OmniObject3D.} We clip the maximum depth value at 10.

\end{document}